\documentclass{article}

\PassOptionsToPackage{numbers, compress}{natbib}
\usepackage[preprint]{neurips_2024}

\usepackage[utf8]{inputenc} 
\usepackage[T1]{fontenc}    
\usepackage{hyperref}       
\usepackage{url}            
\usepackage{booktabs}       
\usepackage{amsfonts}       
\usepackage{nicefrac}       
\usepackage{microtype}      
\usepackage{xcolor}         

\usepackage{enumitem}
\usepackage{bm}
\usepackage{bbm}
\usepackage{amsmath}
\usepackage{wrapfig}
\usepackage{graphicx}
\usepackage[algo2e]{algorithm2e} 
\usepackage{algorithm, algorithmic}

\DeclareFixedFont{\ttb}{T1}{txtt}{bx}{n}{7} 
\DeclareFixedFont{\ttm}{T1}{txtt}{m}{n}{7}  

\usepackage{color}
\definecolor{deepblue}{rgb}{0,0,0.5}
\definecolor{deepred}{rgb}{0.6,0,0}
\definecolor{deepgreen}{rgb}{0,0.5,0}

\usepackage{listings}

\newcommand\pythonstyle{\lstset{
language=Python,
basicstyle=\ttm,
commentstyle=\ttm\color{deepred},
morekeywords={self},              
keywordstyle=\ttb\color{deepgreen},
emph={acquisition_function,
acquisition_function_v0,
acquisition_function_v1,
acquisition_function_v2},
emphstyle=\ttb\color{deepblue},    
stringstyle=\color{deepgreen},
frame=tb,                         
showstringspaces=false,
}}

\lstnewenvironment{python}[1][]
{
\pythonstyle
\lstset{#1}
}
{}

\makeatletter
\newcommand\fs@nocaptionruled{
  \let\@fs@capt\relax
  \def\@fs@pre{}
  \def\@fs@post{\kern2pt\hrule\relax}%
  \def\@fs@mid{\kern2pt\hrule\kern2pt}%
  \let\@fs@iftopcapt\iftrue}
\makeatother

\setlist[itemize]{leftmargin=*}
\usepackage{xspace}
\floatstyle{nocaptionruled}
\restylefloat{algorithm}

\newcommand{\acro}[1]{\textsc{#1}\xspace}
\newcommand{\acronospace}[1]{\textsc{#1}}
\newcommand{\bo}{\acro{bo}}

\newcommand{\af}{\acro{af}}
\newcommand{\gp}{\acro{gp}}
\newcommand{\ei}{\acro{ei}}
\newcommand{\ucb}{\acro{ucb}}
\newcommand{\pofi}{\acronospace{p}of\acro{i}}
\newcommand{\llm}{\acro{llm}}
\newcommand{\mean}{\acro{mean}}
\newcommand{\hpo}{\acro{hpo}}
\newcommand{\svm}{\acro{svm}}
\newcommand{\adaboost}{\acronospace{a}da\acronospace{b}\text{oost}\xspace}
\newcommand{\funbo}{\acronospace{f}\text{un}\acro{bo}}
\newcommand{\funsearch}{\acronospace{f}\text{un}\acronospace{s}\text{earch}\xspace}
\newcommand{\metabo}{\acronospace{m}\text{eta}\bo}
\newcommand{\fsaf}{\acro{fsaf}}
\newcommand{\sty}{Styblinski-Tang\xspace}
\newcommand{\wei}{Weierstrass\xspace}
\newcommand{\gprice}{Goldstein-Price\xspace}
\newcommand{\hart}{Hartmann\xspace}
\newcommand{\expacrosssynthetic}{\acro{ood}-\acronospace{b}ench\xspace}
\newcommand{\expwithinsynthetic}{\acro{id}-\acronospace{b}ench\xspace}
\newcommand{\expwithinhpo}{\acro{hpo}-\acro{id}}
\newcommand{\expwithingps}{\gp{s}-\acro{id}}
\newcommand{\fewshot}{\acro{few}-\acro{shot}}

\DeclareBoldMathCommand{\x}{x}
\DeclareBoldMathCommand{\X}{X}
\DeclareMathOperator*{\argmin}{arg\,min}

\newcommand{\D}{\mathcal{D}}
\newcommand{\Dt}{\mathcal{D}_t}
\newcommand{\eval}{e}
\newcommand{\evolve}{h}
\newcommand{\score}{s}
\newcommand{\fall}{\mathcal{G}}
\newcommand{\ftrain}{\fall_{\acronospace{T}\text{r}}}
\newcommand{\fvalidate}{\fall_{\acro{V}}}
\newcommand{\ftest}{\mathcal{F}}
\newcommand{\Tau}{\mathcal{T}}
\newcommand{\Nislands}{N_{\acro{db}}}
\newcommand{\db}{\acro{db}}
\newcommand{\resultfunbo}{\alpha_{\funbo}}
\newcommand{\inputdim}{d}
\newcommand{\sizesobol}{N_{\acro{sg}}}

\title{FunBO: Discovering Acquisition Functions for Bayesian Optimization with FunSearch}

\author{%
  \textbf{Virginia Aglietti}\thanks{Corresponding author, \texttt{aglietti@google.com}.}\;, \; \textbf{Ira Ktena, \; Jessica Schrouff, \; Eleni Sgouritsa}, \vspace{0.1cm}\\ \; \textbf{Francisco J. R. Ruiz, \; Alan Malek, \; Alexis Bellot, \;  Silvia Chiappa} \vspace{0.3cm}\\
  Google DeepMind\\
}

\begin{document}

\maketitle

\begin{abstract}
The sample efficiency of Bayesian optimization algorithms depends on carefully crafted acquisition functions (\af{s}) guiding the sequential collection of function evaluations. The best-performing \af can vary significantly across optimization problems, often requiring ad-hoc and problem-specific choices. This work tackles the challenge of designing novel \af{s} that perform well across a variety of experimental settings. Based on \funsearch, a recent work using Large Language Models (\llm{s}) for discovery in mathematical sciences, we propose \funbo, an \llm-based method that can be used to learn new \af{s} written in computer code by leveraging access to a limited number of evaluations for a set of objective functions. We provide the analytic expression of all discovered \af{s} and evaluate them on various global optimization benchmarks and hyperparameter optimization tasks. We show how \funbo identifies \af{s} that generalize well in \textit{and} out of the training distribution of functions, thus outperforming established general-purpose \af{s} and achieving competitive performance against \af{s} that are customized to specific function types and are learned via transfer-learning algorithms.
\end{abstract}

\section{Introduction}\label{sec:intro}
Bayesian optimization (\bo) \citep{jones1998efficient, mockus1974bayesian} is a powerful methodology for optimizing complex and expensive-to-evaluate black-box functions which emerge in many scientific disciplines. \bo has been used across a large variety of applications ranging from hyperparameter tuning in machine learning \cite{bergstra2011algorithms, snoek2012practical, cho2020basic} to designing policies in robotics \cite{calandra2016bayesian} and recommending new molecules in drug design \cite{korovina2020chembo}. Two main components lie at the heart of any \bo algorithm: a surrogate model and an acquisition function (\af). The surrogate model expresses assumptions about the objective function, e.g., its smoothness, and it is often given by a Gaussian Process (\gp) \citep{rasmussen2010gaussian}. Based on the surrogate model, the \af determines the sequential collection of function evaluations by assigning a score to potential observation locations. \bo's success heavily depends on the \af's ability to efficiently balance exploitation (i.e. assigning a high score to locations that are likely to yield optimal function values) and exploration (i.e. assigning a high score to regions with higher uncertainty about the objective function in order to inform future decisions), thus leading to the identification of the optimum with the minimum number of evaluations.

Existing \af{s} aim to provide either general-purpose optimization strategies or approaches tailored to specific objective types. For example, Expected Improvement (\ei) \citep{mockus1974bayesian}, Upper Confidence Bound (\ucb) \citep{lai1985asymptotically} and Probability of Improvement (\pofi) \citep{kushner1964new} are all widely adopted \textit{general-purpose} \af{s} that can be used out-of-the-box across \bo algorithms and objective functions. The performance of these \af{s} varies significantly across different types of black-box functions, making the \af choice an ad-hoc, empirically driven, decision. There exists an extensive literature on alternative \af{s} outperforming \ei, \ucb and \pofi, for instance entropy-based \cite{wang2017max} or knowledge-gradient \cite{frazier2008knowledge} optimizers, see \citet[Chapter~7]{garnett2023bayesian} for a review. However, these are generally hard to implement and expensive to evaluate, partly defeating the purpose of replacing the expensive original optimization with the optimization of a much cheaper and faster to evaluate \af.
Other prior works \cite{hsieh2021reinforced,volpp2019meta, wistuba2021few} have instead proposed learning new \af{s} tailored to specific objectives by transferring information from a set of related functions with a given training distribution via, e.g., reinforcement learning or transformers. While such learned \af{s} can outperform general-purpose \af{s}, their generalization performance to objectives outside of the training distribution is often poor (see experimental section and discussion on generalization behaviour in \citet{volpp2019meta}). Defining methodologies that automatically identify new \af{s} capable of outperforming general-purpose and function-specific alternatives, both in and out of the training distribution, remains a significant and unaddressed challenge.

\textbf{Contributions.} We tackle this challenge by formulating the problem of learning novel \af{s} as an algorithm discovery problem and address it by extending \funsearch \cite{romera2023mathematical}, a recently proposed algorithm that uses \llm{s} to solve open problems in mathematical sciences. In particular, we introduce \funbo, a novel method that explores the space of \af{s} written in computer code. \funbo takes an initial \af as input and, with a limited number of evaluations for a set of objective functions, iteratively modifies the \af to improve the performance of the resulting \bo algorithm. Unlike existing algorithms, \funbo outputs code snippets corresponding to improved \af{s}, which can be inspected to (i) identify differences with respect to known \af{s} and (ii) investigate the reasons for their observed performance, thereby enforcing \textit{interpretability}, and (iii) be easily deployed in practice without additional infrastructure overhead. We extensively test \funbo on a range of optimization problems including standard global optimization benchmarks and hyperparameter optimization (\hpo) tasks. For each experiment, we report the explicit functional form of the discovered \af{s} and show that they generalize well to the optimization of functions both in and out of the training distribution, outperforming general-purpose \af{s} while comparing favorably to function-specific ones. To the best of our knowledge, this is the first work exploring \af{s} represented in computer code, thus demonstrating a novel approach to harness the power of \llm{s} for sampling policy design.

\section{Preliminaries}\label{sec:preliminaries}
We consider an expensive-to-evaluate black-box function $f: \mathcal{X} \to \mathbb{R}$ over the input space $\mathcal{X} \subseteq \mathbb{R}^\inputdim$ for which we aim at identifying the global minimum $\x^* = \argmin_{\x \in \mathcal{X}} f(\x)$. We assume access to a set of auxiliary black-box and expensive-to-evaluate objective functions, $\fall = \{g_j\}_{j=1}^J$, with $g_j: \mathcal{X}_j \to \mathbb{R},  \mathcal{X}_j\subseteq \mathbb{R}^{\inputdim_j}$ for $j = 1, \dots, J$, from which we can obtain a set of evaluations.

\textbf{Bayesian optimization.}
\bo seeks to identify $\x^*$ with the smallest number $T$ of sequential evaluations of $f$ given $N$ initial observations $\D=\{\x_i, y_i\}_{i=1}^N$, with $y_i = f(\x_i)$.\footnote{We focus on noiseless observations but the method can be equivalently applied to noisy outcomes.} \bo relies on a probabilistic surrogate model for $f$ which in this work is set to a \gp with prior distribution over any batch of input points $\X = \{\x_1, \dots, \x_N\}$ given by $p(f|\X) = \mathcal{N}(m(\X), K_\theta(\X, \X'))$ with prior mean $m(\X)$ and kernel $K_{\theta}(\X, \X')$ with hyperparameters $\theta$. The posterior distribution $p(f|\D)$ is available in closed form via standard \gp updates. At every step $t$ in the optimization process, \bo selects the next evaluation location by optimizing an \af $\alpha(\cdot|\Dt): \mathcal{X} \to \mathbb{R}$, given the current posterior distribution $p(f|\Dt)$, with $\Dt$ denoting the function evaluations collected up to trial $t$ (including $\D$).  A commonly used \af is the Expected Improvement (\ei), which is defined as $\alpha_{\ei}(\x|\Dt) = (y^* - m(\x|\Dt))\Phi(z) + \sigma(\x|\Dt)\phi(z)$, where $y^*$ denotes the best function value observed in $\Dt$, also called incumbent, $z = (y^* - m(\x|\Dt))/ \sigma(\x|\Dt)$, $\phi$ and $\Phi$ are the standard Normal density and distribution functions, and $m(\x|\Dt)$ and $\sigma(\x|\Dt)$ are the \gp posterior mean and standard deviation computed at $\x \in \mathcal{X}$. Other general-purpose \af{s} proposed in the literature are: \ucb ($\alpha_{\ucb}(\x|\Dt) = m(\x|\Dt) - \beta \sigma(\x|\Dt)$ with hyperparameter $\beta$), \pofi ($\alpha_{\text{\pofi}}(\x|\Dt) = \Phi((y^* - m(\x|\Dt))/\sigma(\x|\Dt))$) and the posterior mean $\alpha_{\text{\mean}}(\x|\Dt) = m(\x|\Dt)$ (denoted by \mean hereinafter).\footnote{We focus on \af{s} that can be evaluated in closed form given the posterior parameters of a \gp surrogate model and exclude those whose computation involve approximations, e.g., Monte-Carlo sampling.}

Unlike general-purpose \af{s}, several works have proposed increasing the efficiency of \bo for a specific optimization problem, say the optimization of $f$, by \emph{learning} problem-specific \af{s} \cite{hsieh2021reinforced,volpp2019meta, wistuba2021few}. These \af{s} are trained on the set $\fall$, whose functions are assumed to be drawn from the same distribution or function class associated to $f$, reflecting a meta-learning setup. ``Function class'' here refers to a set of functions with a shared structure and obtained by, e.g.,\ applying scaling and translation transformations to their input and output values or evaluating the loss function of the same machine learning model, e.g., AdaBoost, on different data sets. For instance, \citet{wistuba2018scalable} learns an \af that is a weighted superposition of \ei{s} by exploiting access to a sufficiently large dataset for functions in $\fall$. \citet{volpp2019meta} considered settings where the observations for functions in $\fall$ are limited and proposed \metabo, a reinforcement learning based algorithm that learns a specialized neural \af, i.e., a neural network representing the \af. The neural \af takes as inputs a set of potential locations (with a given $d$), the posterior mean and variance at those points, the trial $t$ and the budget $T$ and is trained using a proximal policy optimization algorithm \cite{schulman2017proximal}. Similarly, \citet{hsieh2021reinforced} proposed \fsaf, an \af obtained via few-shot adaptation of a learned \af using a small number of function instances in $\fall$. Note that, while general-purpose \af{s} are used to optimize objectives across function classes, learned \af{s} aim at achieving high performance for the single function class to which $f$ and $\fall$ belong. 

\begin{figure}
\begin{minipage}{.6\textwidth}
\begin{algorithm}[H]\small
\textbf{Inputs:} $\ftrain$, $\fvalidate$, $\Nislands$, $B$, $\Tau$
\\
\textbf{Setup:} Initialize $\evolve$ (\textit{Left}), $\eval$ (Fig. \ref{fig:eval_function_funbo}-\ref{fig:score_function_funbo}) and \db with $\Nislands$ islands. Assign $\evolve$ to each island.\\
\While{$\tau < \Tau$
}{  1. Sample two programs from \db and create              prompt (Fig. \ref{fig:funbo_prompt})\\
    2. Get a batch of $B$ samples from the \llm \\
    3. For each correct $\evolve^{\tau}$ in the batch compute $\score_{\evolve^{\tau}}(\ftrain)$\\
    4. Add correct $\evolve^{\tau}$ to \db and update it (see Appendix \ref{sec:programs_database_appendix})\\
    5. Update step $\tau = \tau + 1$
    }
\textbf{Output:}
  Return $\evolve$ in \db with score in the top 20th percentile for $\ftrain$ and highest score on $\fvalidate$.
\end{algorithm}
\end{minipage}
\hfill
\begin{minipage}{.4\textwidth}
\centering
\includegraphics[width=0.9\textwidth]{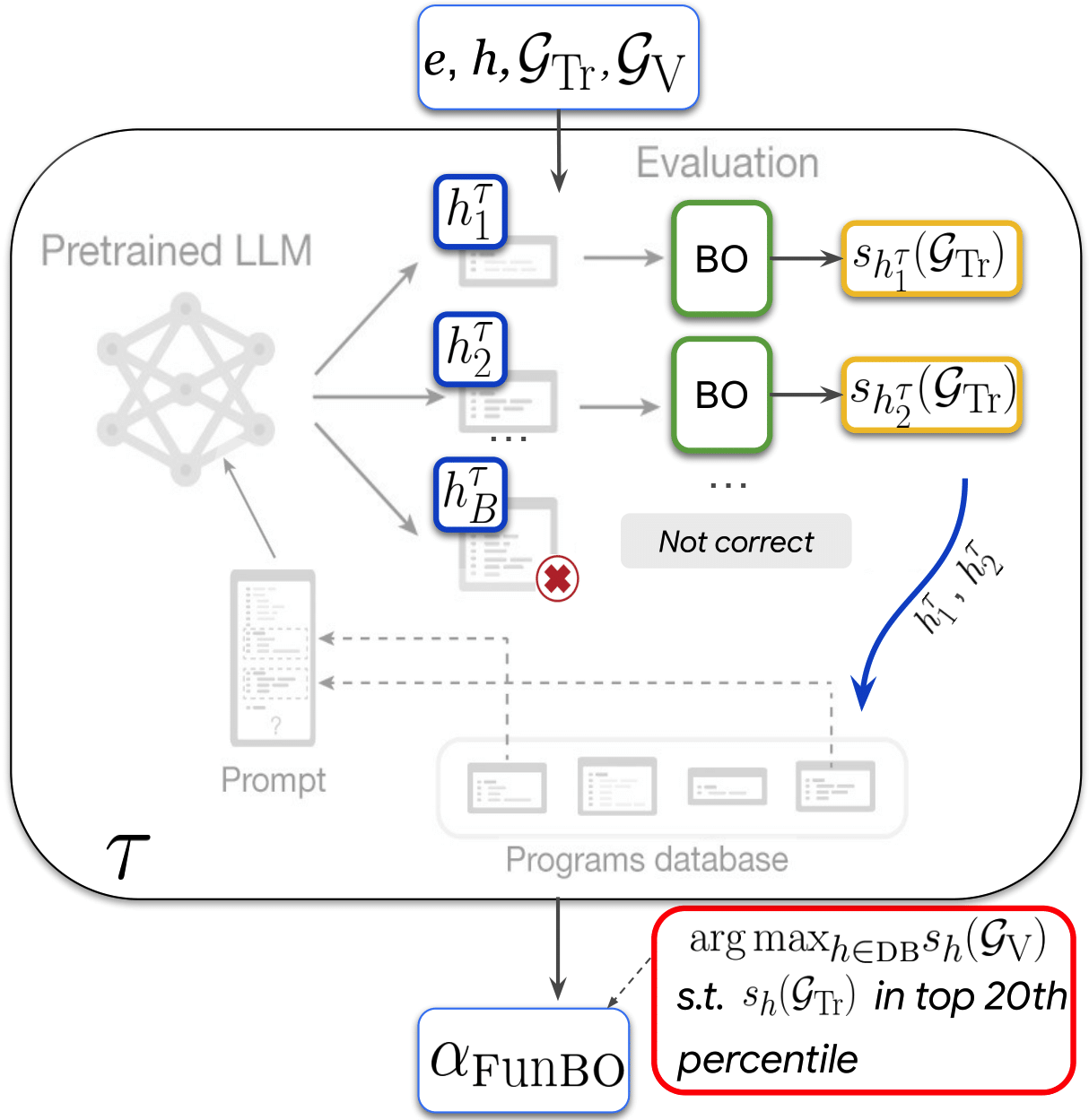}
\end{minipage}
\caption{\textit{Left}: The \funbo algorithm.
\textit{Right}: Graphical representation of \funbo.
The different \funbo component w.r.t. \funsearch  \citep[][Fig. 1]{romera2023mathematical} are highlighted in color.)
}
\label{fig:algorithm_graph}
\vspace{-0.53cm}
\end{figure}

\textbf{\funsearch.} \funsearch \cite{romera2023mathematical} is a recently proposed evolutionary algorithm for \textit{search}ing in the \textit{fun}ctional space by combining a pre-trained \llm used for generating new computer programs with an efficient evaluator, which guards against hallucinations and scores fitness. An example problem that \funsearch tackles is the online bin packing problem~\cite{coffman1984approximation}, where a set of items of various sizes arriving online needs to be packed into the smallest possible number of fixed sized bins. A set of heuristics have been designed for deciding which bin to assign an incoming item to, e.g.,\ ``first fit.'' \funsearch aims at discovering new heuristics that improve on existing ones by taking as inputs: (i) the computer code of an \texttt{evolve} function $\evolve(\cdot)$ representing the initial heuristic to be improved by the \llm, e.g.,\ ``first fit'' and (ii) an \texttt{evaluate} function $\eval(\evolve, \cdot)$, also written in computer code, specifying the problem at hand (also called ``problem specification'') and scoring each $\evolve(\cdot)$ according to a predefined performance metric, e.g.,\ the number of bins used in $\evolve(\cdot)$. The inputs of both $\evolve(\cdot)$ (denoted by $\evolve$ hereinafter) and $\eval(\evolve, \cdot)$ (denoted by $\eval$ hereinafter), are problem specific. A description of $\evolve$'s inputs is provided in the function's docstring\footnote{We focus on Python programs.} together with an explanation of how the function itself is used within $\eval$. Given these initial components, \funsearch prompts an \llm to propose an improved $\evolve$, scores the proposals on a set of inputs, e.g.,\ on different bin-packing instances, and adds them to a programs database. The programs database stores correct $\evolve$ functions\footnote{The definition of a correct function is also problem specific. For instance, a program can be considered correct if it compiles.} together with their respective scores. In order to encourage diversity of programs and enable exploration of different solutions, a population-based approach inspired by genetic algorithms \cite{tanese1989distributed} is adopted for the programs database (\db). At a subsequent step, functions in the database are sampled to create a new prompt, \llm's proposals are scored and stored again. The process repeats for $\tau=1, \dots, \Tau$ until a time budget $\Tau$ is reached and the heuristic with the highest score on a set of inputs is returned.

\section{\funbo}\label{sec:methodology}
\funbo is a \funsearch-based method for discovering novel \af{s} that increase \bo efficiency by exploiting the set of auxiliary objectives $\fall$. In particular, \funbo (i) uses the same prompt and \db structure as \funsearch, but (ii) proposes a new problem specification by viewing the learning of \af{s} as a algorithm discovery problem, and (iii) introduces a novel initialization and evaluation pipeline that is used within the \funsearch structure. \funbo does not make assumptions about similarities between $f$ and $\fall$, nor assumes access to a large dataset for each function in $\fall$. Therefore, \funbo can be used to discover both general-purpose and function-specific \af{s} as well as to adapt \af{s} via few-shots.

\textbf{Method overview.} \funbo sequentially prompts an \llm to improve an \textbf{initial \af} expressed in code so as to enhance the performance of the corresponding \bo algorithm when optimizing objectives in $\fall$. At every step $\tau$ of \funbo, an \llm's \textbf{prompt} is created by including the code for two \af instances generated and stored in a \textbf{programs database} (\db) at previous iterations. With this prompt, a number ($B$) of alternative \af{s} are sampled from the \llm and are evaluated based on their average performance on a subset $\ftrain \subseteq \fall$, which acts as training dataset. The \textbf{evaluation} process for an \af, say $\evolve^{\tau}$ at step $\tau$, on $\ftrain$ gives a numeric score $\score_{\evolve^{\tau}}(\ftrain)$ that is used to store programs in \db and sample them for subsequent prompts. The ``process'' of prompt creation, \llm sampling, and \af scoring and storing repeats until time budget $\Tau$ is reached. Out of the top performing\footnote{In this work we consider the programs with score in the top 20th percentile.} \af{s} on $\ftrain$, the algorithm returns the \af performing the best, on average, in the optimization of $\fvalidate = \fall \backslash \ftrain$, which acts as a validation dataset. When no validation functions are used ($\fall = \ftrain$), the  \af with the highest average performance on $\ftrain$ is returned. Each \funbo component highlighted in bold is described below in more details, along with the complete algorithm and graphical representation in Fig. \ref{fig:algorithm_graph}. We denote the \af returned by \funbo as $\resultfunbo$. 

\begin{figure}
\begin{python}
def acquisition_function(predictive_mean, predictive_var, incumbent, beta=1.0):
  """Returns the index of the point to collect ... (Full docstring in Fig. 8).""" 
  z = (incumbent - predictive_mean) / np.sqrt(predictive_var)
  predictive_std = np.sqrt(predictive_var)
  vals = (incumbent - predictive_mean) * stats.norm.cdf(z) + predictive_std * stats.norm.pdf(z)
  return np.argmax(vals)
\end{python}
\begin{python}
"""Improve Bayesian Optimization by discovering a new acquisition function."""

def acquisition_function_v0(predictive_mean, predictive_var, incumbent, beta=1.0):
    """Returns the index of the point to collect ... (Full docstring in Fig. 8)"""
    # Code for lowest-scoring sampled AF.
    return ...
    
def acquisition_function_v1(predictive_mean, predictive_var, incumbent, beta=1.0):
    """Improved version of `acquisition_function_v0`."""
    # Code for highest-scoring sampled AF.
    return ...

def acquisition_function_v2(predictive_mean, predictive_var, incumbent, beta=1.0):
    """Improved version of the previous `acquisition_function`."""
\end{python}
\caption{\textit{Top}: \funbo's initial \af takes the functional form of \ei with inputs given by the posterior parameter of the \gp at a set of potential sample locations, the incumbent and a parameter $\beta=1$. \textit{Bottom}: \funbo prompt includes two previously generated \af{s} which are sampled from \db and are sorted in ascending order based on the score achieved on $\ftrain$. The \llm generates a third \af, {\small\fontfamily{qcr}\selectfont acquisition\_function\_v2}, representing an improved version of the highest scoring program.
}\vspace{-0.54cm}
\label{fig:funbo_prompt}
\end{figure}

\textbf{Initial \af.} \funbo's initial program $\evolve$ determines the input variables that can be used to generate alternative \af{s} while imposing a prior on the programs the \llm will generate at successive steps. We consider \texttt{acquisition\_function} in Fig. \ref{fig:funbo_prompt} (top) which takes the functional form of the \ei and has as inputs the union of the inputs given to \ei, \ucb and \pofi. The \af returns an integer representing the index of the point in a vector of potential locations that should be selected for the next function evaluation. All programs generated by the \llm share the same inputs and output, but vary in their implementation, which defines different optimization strategies, see for instance the \af generated for one of our experiments in Fig.  \ref{fig:across_classes_af} (left). 

\textbf{Prompt.} At every algorithm iteration, a prompt is constructed by sampling two \af{s}, $\evolve_i$ and $\evolve_j$, previously generated and stored in \db. $\evolve_i$ and $\evolve_j$ are sampled from \db in a way that favours higher scoring and shorter programs (see paragraph below for more details) and are sorted in the prompt in ascending order based on their scores $\score_{\evolve_i}(\ftrain)$ and $\score_{\evolve_j}(\ftrain)$, see the prompt skeleton\footnote{Note that, when $\tau=1$, only the initial program will be available in \db thus the prompt in Fig. \ref{fig:funbo_prompt} will be simplified by removing \texttt{acquisition\_function\_v1} and replacing \texttt{v\_2} with \texttt{v\_1}.} in Fig. \ref{fig:funbo_prompt} (bottom). The \llm is then asked to generate a new \af representing an improved version of the last, higher scoring, program.

\textbf{Evaluation.} 
As expected, the evaluation protocol is critical for the discovery of appropriate \af{s}. Our novel evaluation setup, unlike the one used in \funsearch, entails performing a full \bo loop to evaluate program fitness. In particular, each function generated by the \llm is (i) checked to verify it is correct, i.e., it compiles and returns a numerical output; (ii) scored based on the average performance of a \bo algorithm using $\evolve^{\tau}$ as an \af on $\ftrain$. Evaluation is performed by running a full \bo loop with $\evolve^{\tau}$ for each function $g_j \in \ftrain$ and computing a score that contains two terms: a term that rewards \af{s} finding values close to the true optimum, and a term that rewards \af{s} finding the optimum in fewer evaluations (often called trials). Specifically, we use the score:
\begin{align}
   \score_{\evolve^{\tau}}(\ftrain) = \frac{1}{|\ftrain|}\sum_{j=1}^J \left[ \left(1 - \frac{g_{j}(\x^*_{j, \evolve^{\tau}}) - y^*_j}{g_j(\x^{t=0}_j) -  y^*_j}\right) + \left(1 - \frac{T_{\evolve^{\tau}}}{T}\right)\right]
    \label{eq:score}
\end{align}
where, for each $g_j$, $y^*_j$ is the known true optimum, $\x^{t=0}_j$ gives the optimal input value at $t=0$ which is assumed to be different from the true one, $\x^*_{j, \evolve^{\tau}}$ is the found optimal input value with $\evolve^{\tau}$ and $T_{\evolve^{\tau}}$ gives the number of trials out of $T$ that $\evolve^{\tau}$ selected before reaching $y^*_j$ (if the optimum was not found, then $T_{\evolve^{\tau}}=T$ to indicate that all available trials have been used). The first term in the square brackets of Eq. \eqref{eq:score} quantifies the discrepancy between the function values at the returned optimum and the true optimum. This term becomes zero when $\x^*_{j, \evolve^{\tau}}$ equals $\x^{t=0}_j$, indicating a failure to explore the search space. Conversely, if $\evolve^{\tau}$ successfully identifies the true optimum, such that $g_j(\x^*_{j, \evolve^{\tau}}) = y^*_j$, this term reaches its maximum value of one. The second term in Eq. \eqref{eq:score} captures how quickly $\evolve^{\tau}$ identifies $y^*_j$. When $T_{\evolve^{\tau}} = T$, indicating the algorithm has not converged, this term becomes zero, and the score is solely determined by the discrepancy between the discovered and true optimum. If, instead, the algorithm reaches the global optimum, this term represents the proportion of trials, out of the total budget $T$, needed to do so. Code for the evaluation process is presented in Appendix \ref{sec:pseudo_code}.

\textbf{Programs database.} Similar to \funsearch, scored \af{s} are added to \db, which keeps a population of correct programs following an island model \cite{tanese1989distributed}. \db is initialized with a number $N_{\db}$ of islands that evolve independently. Sampling of $\evolve_i$ and $\evolve_j$ from \db is done by first uniformly sampling an island and, within that island, sampling programs by favouring those that are shorter and higher scoring. A new program generated when using $\evolve_i$ and $\evolve_j$ in the prompt is added to the same island and, within that, to a cluster of programs performing similarly on $\ftrain$, see Appendix \ref{sec:programs_database_appendix} for more details.

\section{Experiments}
\begin{figure}
\centering
\begin{minipage}{.5\textwidth}
\begin{python}
def acquisition_function(predictive_mean
    predictive_var, incumbent, beta=1.0):
  """Returns the index of the point to collect...""" 
  predictive_std = np.sqrt(predictive_var)
  diff_mean_std = (incumbent - predictive_mean
  + beta * predictive_std)
  z = diff_mean_std / predictive_std
  vals = (diff_mean_std * stats.norm.cdf(z)
      + predictive_std * stats.norm.pdf(z))
  return np.argmax(vals)
\end{python}
\end{minipage}%
\begin{minipage}{.46\textwidth}
    \centering
    \includegraphics[width=1.0\textwidth]{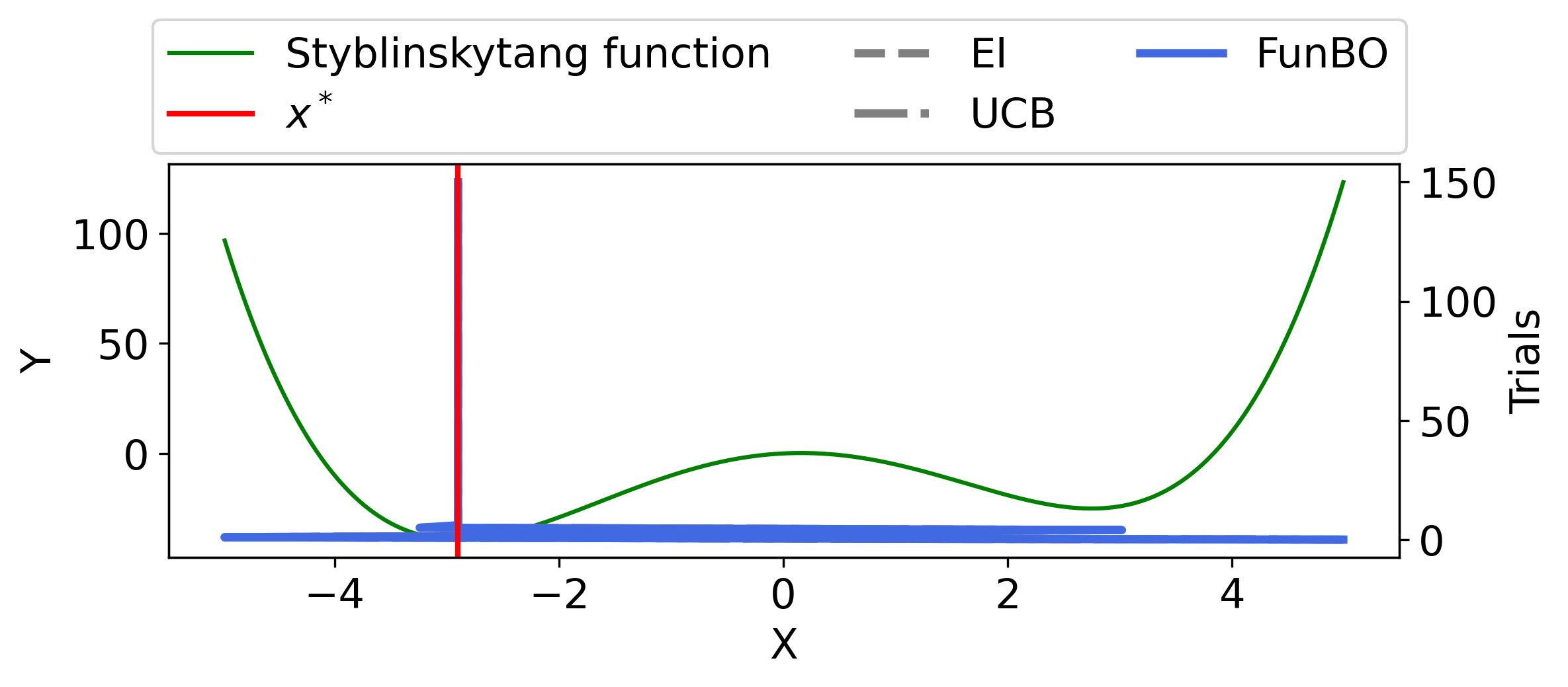}
    \includegraphics[width=0.98\textwidth]{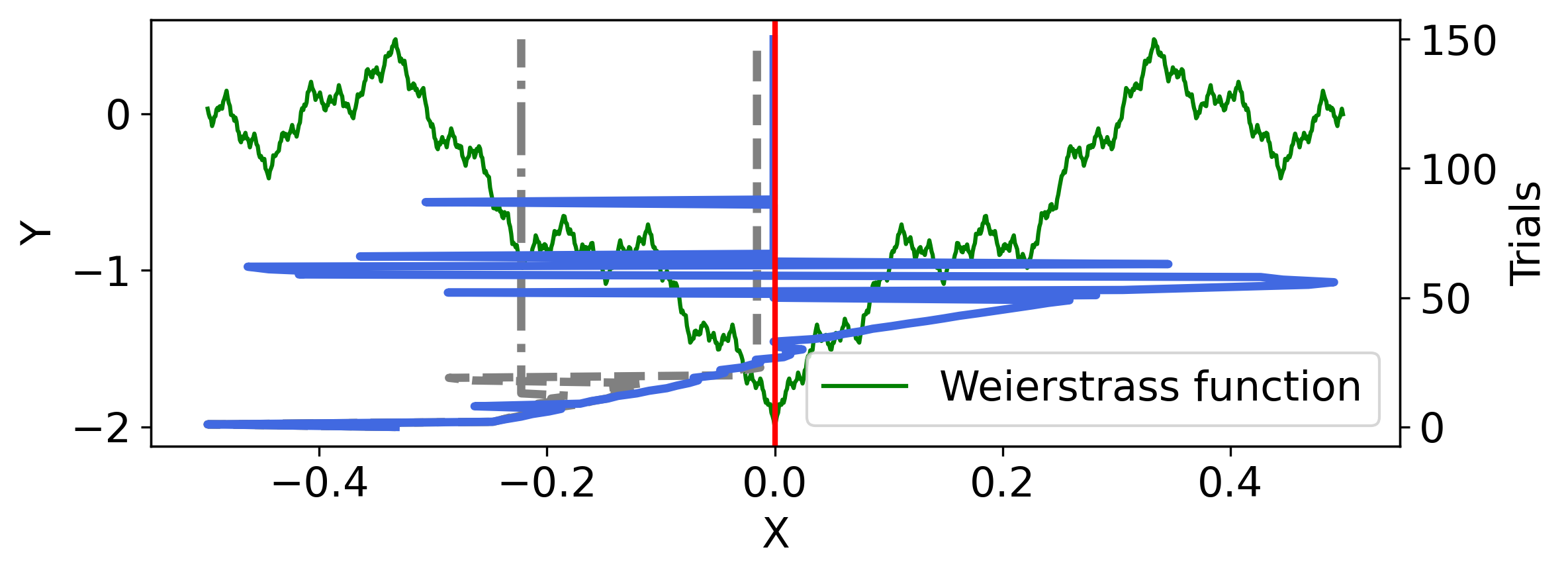}
\end{minipage}
\caption{\expacrosssynthetic. \textit{Left:} Code for $\resultfunbo$. \textit{Right}: Different \af{s} trading-off exploration and exploitation for two one-dimensional objective functions (green lines). Blue and gray trajectories track the points queried by $\resultfunbo$, \ei and \ucb over 150 steps (right $y$-axis). All \af{s} behave similarly for \sty (top, note that trajectories are overlapping), converging to the true optimizer (red vertical line) in fewer than 25 trials. Instead, for \wei (bottom), \ei and \ucb get stuck after a few trials while $\resultfunbo$ continues to explore, eventually converging to the ground truth optimum.}
\label{fig:across_classes_af}
\end{figure}

Our experiments explore \funbo's ability to generate novel and efficient \af{s} across a wide variety of settings. In particular, we demonstrate its potential to generate \af{s} that generalize well to the optimization of functions both in distribution (\acro{id}, i.e.\ within function classes) and out of distribution (\acro{ood}, i.e.\ across function classes) by running three different types of experiments:
\begin{enumerate}[leftmargin=*]
\item \expacrosssynthetic tests generalization across function classes by running \funbo with $\fall$ containing different standard global optimization benchmarks and testing on a set $\ftest$ that similarly comprises diverse functions in terms of smoothness, input ranges and dimensionality and output magnitudes. We do not scale the output values nor normalise the input domains to facilitate learning, but rather use the objective functions as available in standard \bo packages out-of-the-box. In this case $\fall$ and $\ftest$ do not share any particular structure, thus the generated \af{s} are closer to general-purpose \af{s}.
\item \expwithinsynthetic, \expwithinhpo and \expwithingps test \funbo-generated \af{s} within function classes for standard global optimization benchmarks, \hpo tasks, and general function classes, respectively. As this setting is closer to the one considered by meta-learning approaches introduced in Section \ref{sec:preliminaries}, we compare \funbo against \metabo \cite{volpp2019meta},\footnote{We used the author-provided implementation at \texttt{https://github.com/boschresearch/MetaBO}.} the state-of-the-art transfer acquisition function. 
\item \fewshot demonstrates how \funbo can be used in the context of few-shot fast adaptation of an \af. In this case, the \af is learnt using a general function class as $\fall$ and is then tuned, using a very small (5) number of examples, to optimize a specific synthetic function. We compare our approach to \citet{hsieh2021reinforced},\footnote{We used the author-provided implementation at \texttt{https://github.com/pinghsieh/FSAF}.} the most relevant few-shot learning method. 
\end{enumerate}

\begin{wrapfigure}[16]{r}{0.5\textwidth}
\vspace{-0.4cm}
\centering
\includegraphics[width=0.5\textwidth]{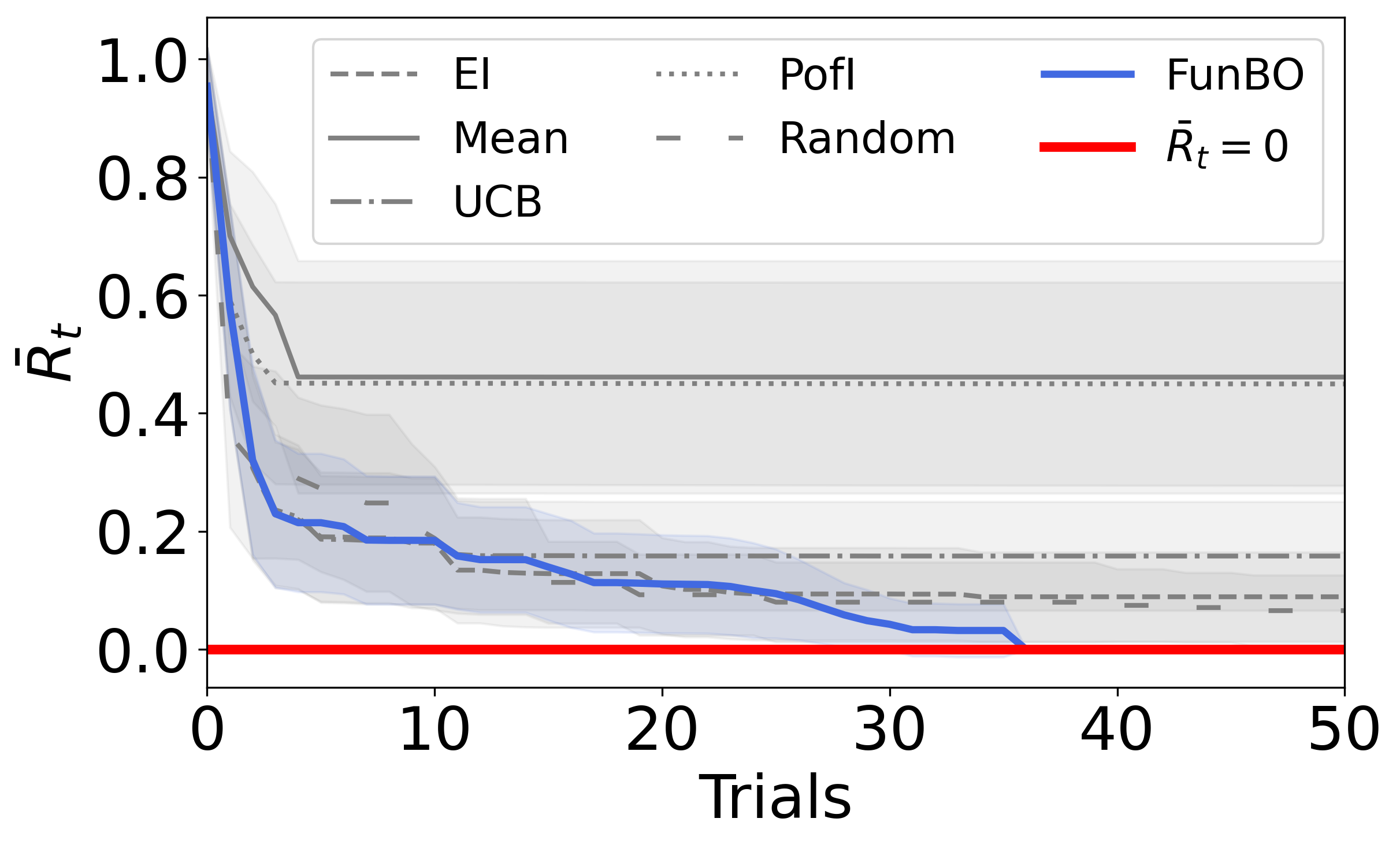}
\vspace{-0.6cm}
\caption{\expacrosssynthetic. Average \bo performance when using known general purpose \af{s} and $\resultfunbo$. Shaded area gives $\pm$ standard deviations$/2$. The red line gives $\bar{R}_t = 0$, i.e.\ zero average regret.}
\label{fig:across_classes_performance}
\end{wrapfigure}

We report all results in terms of normalized average simple regret on a test set, $\bar{R}_t$, as a function of the trial $t$. For an objective function $f$, this is defined as $R_t = f(\x^*_t) - y^*$ where $y^*$ is the true optimum and $\x^*_t$ is the best selected point within the data collected up to $t$. As $\ftest$ might include functions with different scales, we normalize the regret values to be in $[0, 1]$ before averaging them. To isolate the effects of different acquisition functions, we employ the same setting across all methods in terms of (i) number of trials $T$, (ii) hyperparameters of the \gp surrogate models (tuned offline), (iii) evaluation grid for the \af, which is set to be a Sobol grid \cite{sobol1967distribution} on the input space, and (iv) initial design, which includes the input point giving the maximum function value on the grid. All experiments are conducted using \funsearch with default hyperparameters in \citet{romera2023mathematical}\footnote{See code at \texttt{https://github.com/google-deepmind/funsearch}.} unless otherwise stated. We employ Codey, an \llm fine-tuned on a large code corpus and based on the PaLM model family \cite{google2023palm2}, to generate \af{s}.\footnote{Codey is publicly accessible via its \acro{api} \cite{codeyapi}. For \af sampling, we used 5 Codey instances running on tensor processing units on a computing cluster. For scoring, we used 100 \acro{cpu}{s} evaluators per \llm instance.
}

\textbf{\expacrosssynthetic.} We test the capabilities of \funbo to generate an \af that performs well \emph{across} function classes by including the one-dimensional functions Ackley, Levy, and Schwefel in $\ftrain$ and using the one-dimensional Rosenbrock function for $\fvalidate$. We test the resulting $\resultfunbo$ on nine very different objective functions: Sphere ($\inputdim=1$), \sty ($\inputdim=1$), \wei ($\inputdim=1$), Beale ($\inputdim=2$), Branin ($\inputdim=2$), Michalewicz ($\inputdim=2$), \gprice ($\inputdim=2$) and \hart with both $\inputdim=3$ and $\inputdim=6$. We do not compare against \metabo as (i) it was developed for settings in which the functions in $\fall$ and $\ftest$ belong to the same class and, (ii) the neural \af is trained with evaluation points of a given dimension, thus it cannot be deployed for the optimization of functions across different $\inputdim$. For completeness, we report a comparison with a dimensionality-agnostic version of \metabo in Appendix \ref{sec:across_class_experiments} (Fig. \ref{fig:across_classes_performance_appendix}) together with all experimental details, e.g.,\ input ranges and hyperparameter settings.

\textit{\af interpretation:} In this experiment, $\resultfunbo$ (Fig. \ref{fig:across_classes_af}, left) represents a combination of \ei and \ucb which, due to the  \texttt{beta*predictive\_std} term, is more exploratory than \ei but, considering the incumbent value, still factors in the expected magnitude of the improvement and reduces to \ei when \texttt{beta=0}. This determines the way $\resultfunbo$ trades-off exploration and exploitation which can be visualized by looking at the "exploration path", i.e., the sequence of $\x$ values selected over $t$, as shown in the right plots of Fig. \ref{fig:across_classes_af} ($t$ measured on the secondary y-axis). For objective functions that are smooth, for example \sty (top plot), the exploration path of $\resultfunbo$ matches those of \ei and \ucb. In this scenario, all \af{s} exhibit similar behavior, converging to $\x^*$ (red vertical line) with less than 25 trials. When instead the objective function has a lot of local optima (bottom plot) as in \wei, both \ei and \ucb get stuck after a few trials while \funbo keeps on exploring the search space eventually converging to $\x^*$. Notice how in this plot the convergence paths of all \af{s} differ and only the blue line aligns with the red line, i.e., converges to $\x^*$, after a few trials. 

Using $\resultfunbo$ to optimize the nine functions in $\ftest$ leads to a fast and accurate convergence to the global optima (Fig. \ref{fig:across_classes_performance}). The same is confirmed when extending the test set to include 50 scaled and translated instances of the functions in $\ftest$ (Fig. \ref{fig:across_classes_performance_appendix}, right).

\begin{figure}
    \centering
    \includegraphics[width=0.325\textwidth]{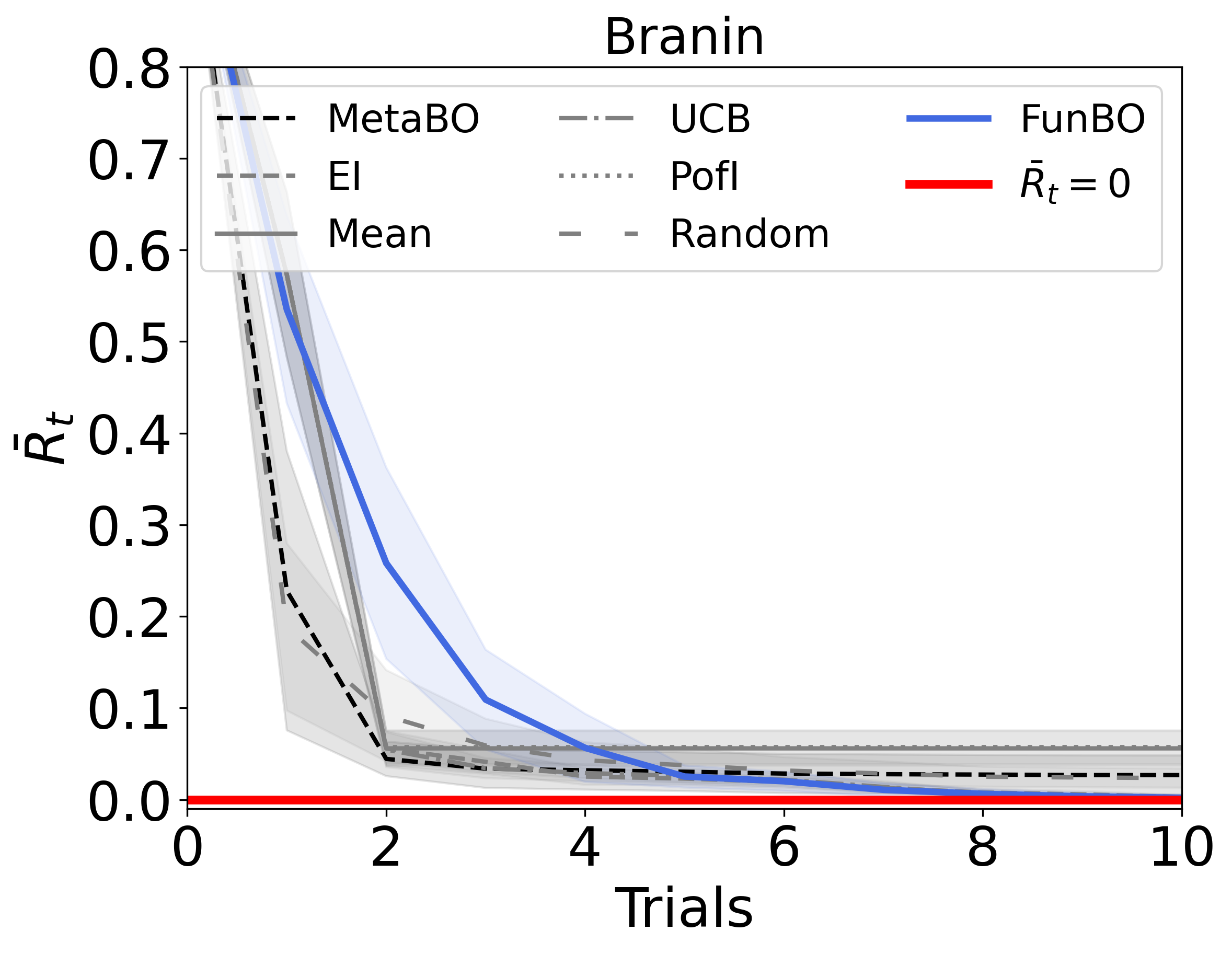}
    \includegraphics[width=0.325\textwidth]{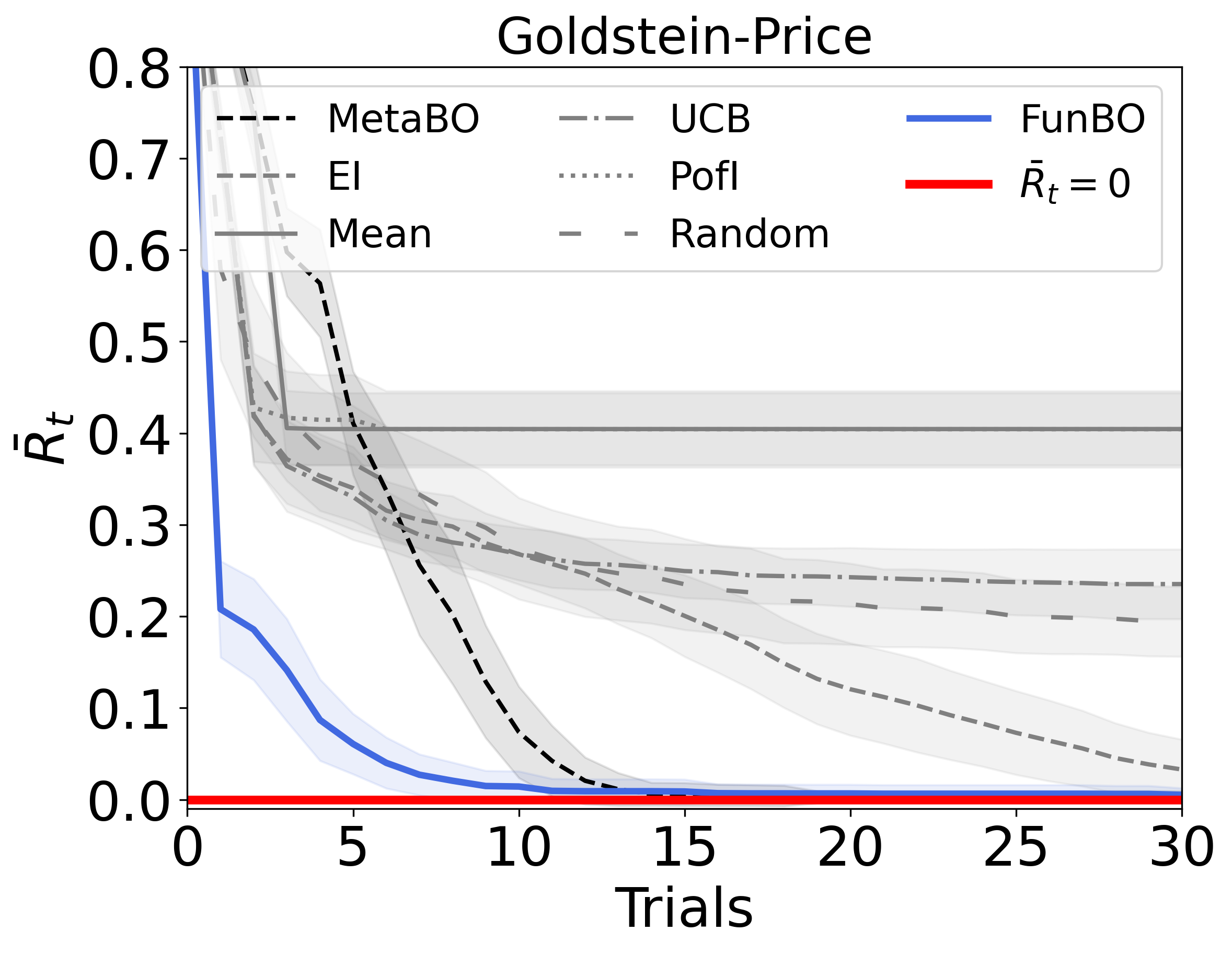}
    \includegraphics[width=0.325\textwidth]{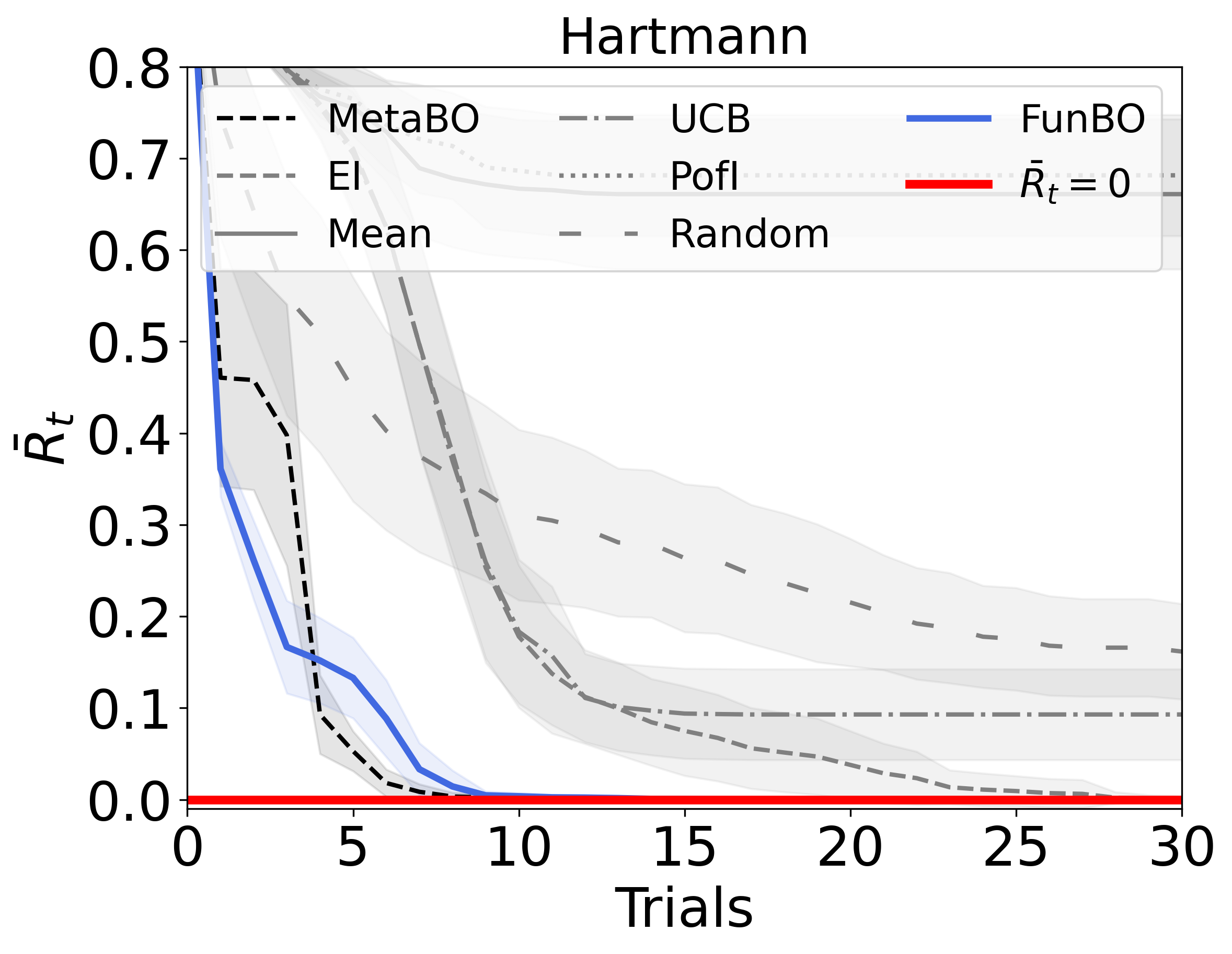}
    \caption{\expwithinsynthetic. Average \bo performance when using known general purpose \af{s} (gray lines), the \af learned by \metabo (black dashed line) and $\resultfunbo$ (blue line) on 100 function instances. Shaded area gives $\pm$ standard deviations$/2$. The red line represents $\bar{R}_t=0$, i.e.\ zero average regret.}
    \label{fig:branin_gprice_hm3}
\end{figure}

\textbf{\expwithinsynthetic.}
Next we evaluate \funbo capabilities to generate \af{s} that perform well \textit{within} function classes using Branin, \gprice and \hart ($\inputdim = 3$). For each of these three functions, we train both \funbo and \metabo with $|\fall| = 25$ instances of the original function obtained by scaling and translating it with values in $[0.9, 1.1]$ and $[-0.1, 0.1]^{\inputdim}$ respectively.\footnote{Throughout the paper we adopt \metabo's translation and scaling ranges.} For \funbo we randomly assign 5 functions in $\fall$ to $\fvalidate$ and keep the rest in $\ftrain$. We test the performance of the learned \af{s} on another 100 instances of the same function, with randomly sampled values of scale and translation from the same ranges. We additionally compare against a \bo algorithm that uses \ei, \ucb, \pofi, \mean or a random selection of points. All hyper-parameter settings for this experiment are provided in Appendix \ref{sec:app_expwithinsynthetic}. Across all objective functions, $\resultfunbo$ leads to a convergence performance that is close to or outperform both general purpose and meta-learned \af{s} (Fig. \ref{fig:branin_gprice_hm3}). The \af{s} found in this experiment (code in Figs. \ref{fig:af_branin}-\ref{fig:af_hm3}) are ``customized'' to a given function class thus being closer, in spirit, to the transfer \af. To further validate the generalizability of $\resultfunbo$ found in \expacrosssynthetic, we tested such \af across instances of Branin, \gprice and \hart (Fig. \ref{fig:branin_gprice_hm3_with1D_appendix}, green line). We found it to perform well against general purpose \af{s} thus confirming the strong results observed in \expacrosssynthetic while being, as expected, slower than \af{s} customized to a specific objective.

\begin{figure}
    \centering
    \includegraphics[width=6cm, height=4.13cm]{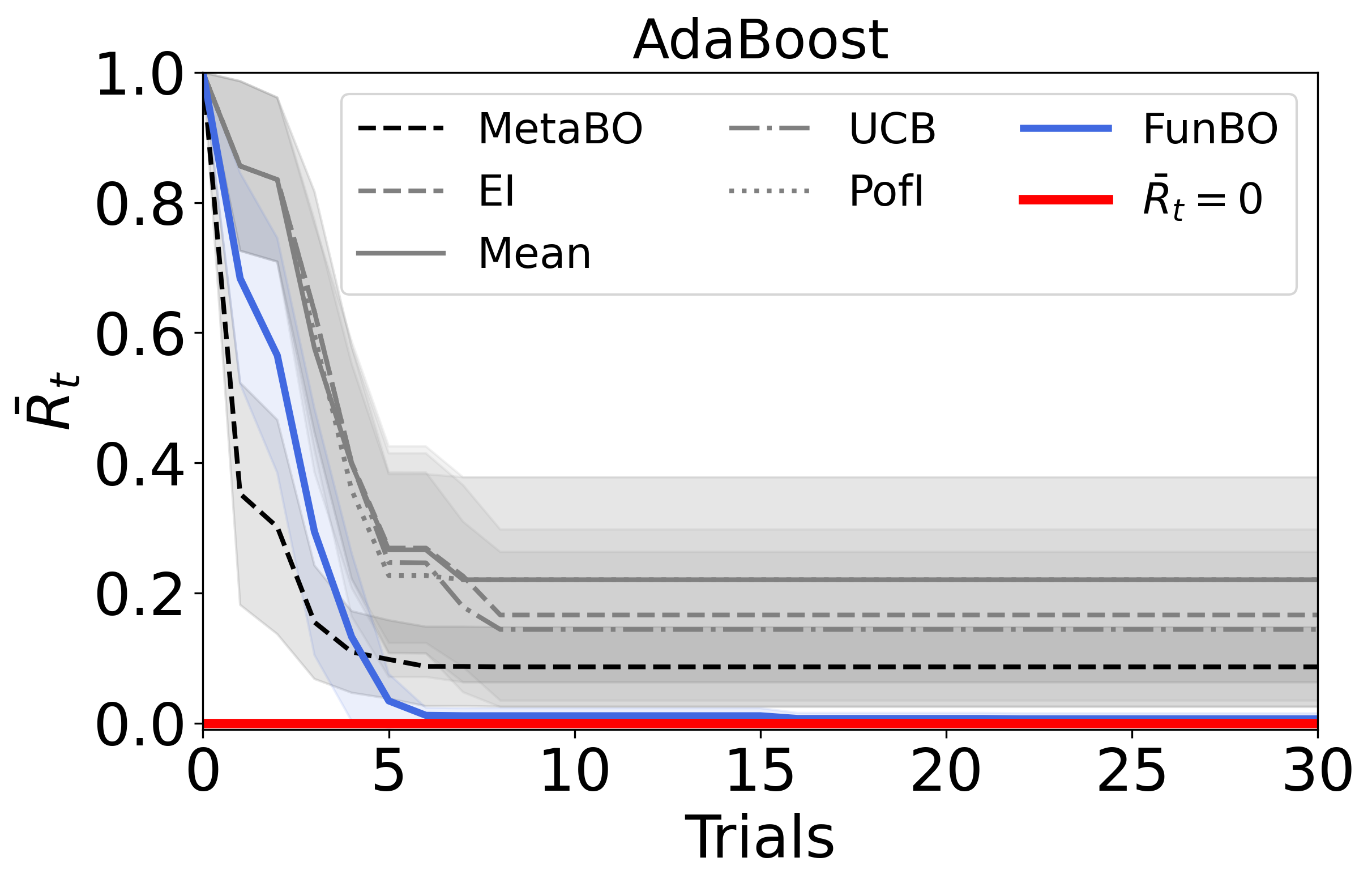}
    \includegraphics[width=6cm, height=5cm, keepaspectratio
    ]{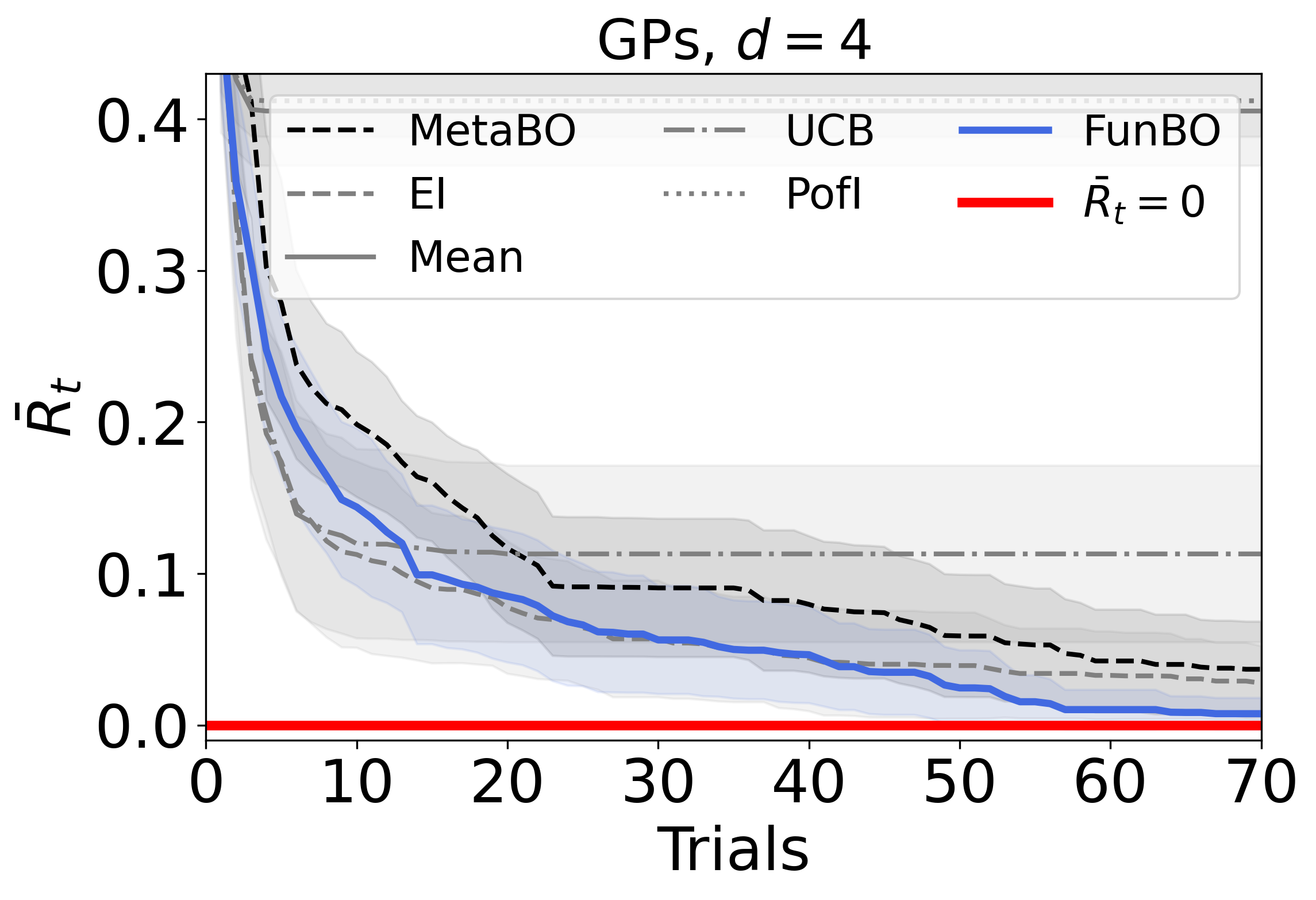}
    \caption{Average \bo performance when using known general purpose \af{s} (gray lines), the \af learned by \metabo (black dashed line) and $\resultfunbo$ (blue line). Shaded area gives $\pm$ standard deviations$/2$. The red line represents $\bar{R}_t=0$, i.e.\ zero average regret. \textit{Left}: \expwithinhpo. \textit{Right}: \expwithingps with $d=4$.}
    \label{fig:adaboost_gps}
\end{figure}

\textbf{\expwithinhpo.} We test \funbo on two \hpo tasks where the goal is to minimize the loss ($\inputdim=2$) of an \acro{rbf}-based \svm and an \adaboost algorithm.\footnote{We use precomputed loss values across 50 datasets given as part of the HyLAP project (\texttt{http://www.hylap.org/}). For \svm, the two hyperparameters are the \acro{rbf} kernel parameter and the penalty parameter while for \adaboost they correspond to the number of product terms and the number of iterations.} As in \expwithinsynthetic, we test the ability to generate \af{s} that generalize well within function classes. Therefore, we train \funbo and \metabo with losses computed on a random selection of 35 of the 50 available datasets and test on losses computed on the remaining 15 datasets. For \funbo we randomly assign 5 dataset to $\fvalidate$ and keep the rest in $\ftrain$. \funbo identifies \af{s} (code in Fig. \ref{fig:af_adaboost}-\ref{fig:af_svm}) that outperform all other \af{s} in \adaboost (Fig. \ref{fig:adaboost_gps}, left) while performing similarly to general purpose or meta-learned \af{s} for \svm (Fig. \ref{fig:adaboost_svm_appendix}). Across the two tasks, $\resultfunbo$ found in \expacrosssynthetic still outperforms general-purpose \af{s} while yielding slightly worse performance compared to \metabo and \funbo customized \af{s} (Fig. \ref{fig:adaboost_svm_appendix}, green lines).

\textbf{\expwithingps.} Similar results are obtained for general function classes whose members do not exhibit any particular shared structure. We let $\ftrain$ include 25 functions sampled from a \gp prior with $\inputdim=3$, \acro{rbf} kernel and length-scale drawn uniformly from $[0.05, 0.5]$. We test the found \af on 100 other \gp samples defined both for $d=3$ and $d=4$ and length-scale values sampled similarly. 
\begin{wrapfigure}[14]{r}{0.5\textwidth}
\vspace{-0.25cm}
\centering
\includegraphics[width=0.5\textwidth]{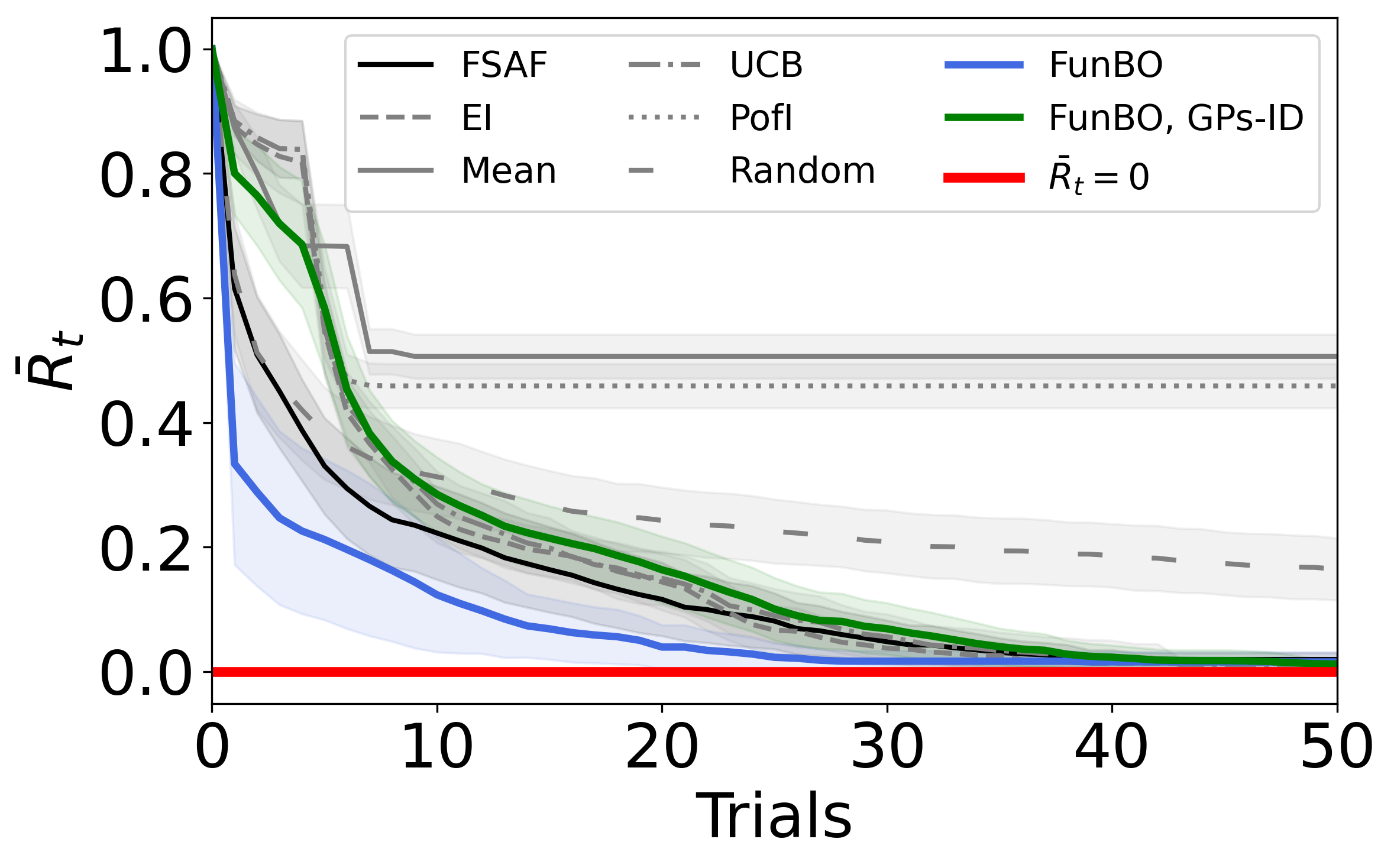}
\vspace{-0.5cm}
\caption{\fewshot.}
\label{fig:few_shots}
\end{wrapfigure}
As done by \cite{volpp2019meta}, we consider a dimensionality-agnostic version of \metabo that allows deploying the function learned from $d=3$ functions on $d=4$ objectives.  We found $\resultfunbo$ to outperform all other \af{s} (code in Fig. \ref{fig:af_gps}) in $d=4$ (Fig. \ref{fig:adaboost_gps}, right) while matching \ei and outperforming \metabo in $d=3$ (Fig. \ref{fig:gps_appendix}, left).

\textbf{\fewshot.} We conclude our experimental analysis by demonstrating how \funbo can be used in the context of few-shot adaptation. In this setting, we aim at learning an \af customized to a specific function class by ``adapting'' an initial \af with a small number of instances from the target class. We consider Ackley ($\inputdim=2$) as the objective function and compare against \fsaf \cite{hsieh2021reinforced}, which is the closest few-shot adaptation method for \bo. \fsaf trains the initial \af with a set of \gp{s}, adapts it using 5 instances of scaled and translated Ackley functions, then tests the adapted \af on 100 additional Ackley instances, generated in the same manner. Note that \fsaf uses a large variety of \gp functions with different kernels and various hyperparameters for training the initial \af. On the contrary, \funbo few-shot adaptation is performed by setting the initial $\evolve$ function to the one found in \expwithingps (Fig. \ref{fig:few_shots}, green line) using 25 \gp{s} with \acro{rbf} kernel, and including  the 5 instances of Ackley used by \fsaf in $\ftrain$. Despite the limited training set, \funbo adapts very quickly to the new function instances, identifying an \af (code in Fig. \ref{fig:af_fewshots}) that outperforms both general purpose \af{s} and \fsaf (Fig. \ref{fig:few_shots}, blue line). 

\section{Related work}
\textbf{\llm{s} as mutation operators.} \funbo expands \funsearch \cite{romera2023mathematical}, an evolutionary algorithm pairing an \llm with an evaluator to solve open problems in mathematics and algorithm design. Prior to \funsearch, the idea of using \llm{s} as mutation operators paired with a scoring mechanism had been explored to a create a self-improvement loop \cite{lehman2023evolution}, to optimize code for robotic simulations, or to evolve stable diffusion images with simple genetic algorithms \cite{meyerson2023language}. Other works explore the use of \llm{s} to search over neural network architectures described with Python code \cite{nasir2023llmatic, zheng2023can, chen2024evoprompting}, find formal proofs for automatic theorem proving \cite{polu2020generative, jiang2022thor} or automatically design heuristics \cite{liu2024example}.

\textbf{Meta-learning for \bo.} Our work is also related to the literature on meta-learning for \bo. In this realm, several studies have focused on meta-learning an accurate surrogate model for the objective function exploiting observations from related functions, for instance by using standard multi-task \gp{s} \cite{swersky2013multi, yogatama2014efficient} or ensembles of \gp models \cite{feurer2018scalable, wistuba2018scalable, wistuba2021few}. Others have focused on meta-learning general purpose optimizers by using recurrent neural networks with access to gradient information \cite{chen2017learning} or transformers \cite{chen2022towards}. More relevant to our work are studies focusing on transferring information from related tasks by learning novel \af{s} that more efficiently solve the classic exploration-exploitation trade-off in \bo algorithms \cite{volpp2019meta, hsieh2021reinforced, maraval2024end}. In contrast to prior works in this literature, \funbo produces \af{s} that are more interpretable, simpler and cheaper to deploy than neural network-based \af{s} and generalize not only within specific function classes but also across different classes.

\textbf{\llm{s} and black-box optimization.} Several works investigated the use of \llm{s} to solve black-box optimization problems. For instance, both \citet{liu2024large} and \citet{yang2023large} framed optimization problems in natural language and asked \llm{s} to iteratively propose promising solutions and/or evaluate them. Similarly, \citet{ramos2023bayesian} replaced surrogate modeling with \llm{s} within a \bo algorithm targeted at catalyst or molecule optimization.  Other works have focused on exploiting black-box methods for prompt optimization \cite{sun2022black, chen2023instructzero, cheng2023black, fernando2023promptbreeder}, solving \hpo tasks with \llm{s} \cite{zhang2023using} or identifying optimal \llm hyperparameter settings via black-box optimization approaches \cite{wang2023cost, tribes2024hyperparameter}. Unlike these approaches, we do not use \llm{s} to replace the entire optimization process, but rather leverage their creative capabilities to enhance a critical component within an existing \bo algorithm, thus narrowing the search space and ensuring interpretability.

\section{Conclusions and Discussion}\label{sec:conclusions_discussion}
We tackled the problem of discovering novel, well performing \af{s} for \bo through \funbo, a \funsearch-based algorithm which explores the space of \af{s} by letting an \llm iteratively modify the \af expression in native computer code to improve the efficiency of the corresponding \bo algorithm. We have shown across a variety of settings that \funbo learns \af{s} that generalize well within and across function classes while being easily adaptable to specific objective functions of interest with only a few training examples.

\textbf{Limitations.} \funbo inherits the strengths of \funsearch along with some of its inherent constraints. While \funsearch allows finding programs that are concise and interpretable, it works best for programs that can be quickly evaluated and for which the score provides an accurate quantification of the improvement achieved. Therefore, a potential limitation of \funbo is the computational overhead associated with running a full \bo loop for each function in $\fall$, which significantly increases the evaluation time of every sampled \af (especially when $T$ is high). This limits the scalability of \funbo for larger sets $\fall$ and hinders its application to more complex optimization problems, such as those with multiple objectives. In addition, the simple metric considered in this work in Eq. \eqref{eq:score}, only captures the distance from the true optimum and the number of trials needed to identify it. More research needs to be done to understand if a metric that better characterizes the convergence path for a given \af can improve \funbo performance. Furthermore, each \funbo experiment shown in this work required obtaining a large number of \llm samples. This means that the overall cost of experiments, which depends on the \llm used as well as the algorithm's implementation (e.g.\ single threaded or distributed, as originally proposed by \funsearch), can be high. Finally, as reported in \citep{romera2023mathematical}, the variance in the quality of the \af found by \funbo is high. This is due to the randomness in both the \llm sampling and the evolutionary procedure. While we were able to reproduce the results shown for \expwithinsynthetic, \expwithinhpo and \expwithingps with different \funbo experiments, finding \af{s} that perform well across function classes required multiple \funbo runs.

\textbf{Future work.} This work opens up several promising avenues for future research. While our focus here was on the simplest single-output \bo algorithm with a \gp surrogate model, \funbo can be extended to learn new \af{s} for various adaptations of this problem, such as constrained optimization, noisy evaluations, or alternative surrogate models. Additionally, \funbo demonstrates the potential to harness the power of \llm{s} while maintaining the interpretability of \af{s} expressed in code. This opens an exciting avenue for exploring how and what assumptions can be encoded within \af{s}, based on the desired program characteristics and prior knowledge about the objective function. Finally, the discovered \af{s} might have intrinsic value, independently on how they were discovered. Future work could focus on more extensively test their properties and identify those that can be added to the standard suite of \af{s} available in \bo packages.
\bibliography{bib}

\newpage
\appendix

\section{Code for \funbo components}\label{sec:pseudo_code}
Fig. \ref{fig:full_initial_af_funbo} gives the Python code for the initial acquisition function used by \funbo, including the full docstring. The docstring describes the inputs of the function and the way in which the function itself is used within the evaluate function $\eval$. Evaluation of the functions generated by \funbo is done by first running a full \bo loop (see Fig. \ref{fig:eval_function_funbo} for Python code) and then, based on its output (the initial optimal input value, the true optimum, the found optimum and the percentage of steps taken before finding the latter), computing the score as in the Python code of Fig. \ref{fig:score_function_funbo}. Note how the latter captures how accurately and quickly a \bo algorithm using the proposed \af finds the true optimum.

\begin{figure}
\begin{python}
import numpy as np
from scipy import stats

def acquisition_function(predictive_mean, predictive_var, incumbent, beta=1.0):
  """Returns the index of the point to collect in a vector of eval points.

  Given the posterior mean and posterior variance of a GP model for the objective function,
  this function computes an heuristic and find its optimum. The next function evaluation
  will be placed at the point corresponding to the selected index in a vector of
  possible eval points.

  Args:
    predictive_mean: an array of shape [num_points, dim] containing the predicted mean
        values for the GP model on the objective function for `num_points` points of
        dimensionality `dim`.
    predictive_var: an array of shape [num_points, dim] containing the predicted variance
        values for the GP model on the objective function for `num_points` points
        of dimensionality `dim`.
    incumbent: current optimum value of objective function observed.
    beta: a possible hyperparameter to construct the heuristic.

  Returns:
    An integer representing the index of the point in the array of shape [num_points, dim]
    that needs to  be selected for function evaluation.
  """
  z = (incumbent - predictive_mean) / np.sqrt(predictive_var)
  predictive_std = np.sqrt(predictive_var)
  vals = (incumbent - predictive_mean) * stats.norm.cdf(z) + predictive_std * stats.norm.pdf(z)
  return np.argmax(vals)
\end{python}
\caption{Python code for \funbo initial $\evolve$ function with full docstring.}
\label{fig:full_initial_af_funbo}
\end{figure}

\begin{figure}
\begin{python}
"""Evaluate an AF with a full BO loop for the objective f."""

import GPy
import numpy as np
import utils

def run_bo(
    f,    # objective function to minimize
    acquisition_function, # h given by LLM
    num_eval_points = 1000,
    num_trials = 30):
  """Run a BO loop and return the minimum objective functions found and the percentage of
  trials required to reach it."""
  
  # Get evaluation points for AF. get_eval_points() returns a given number of points on a
  # Sobol grid on the f's input space
  eval_points = utils.get_eval_points(f, num_eval_points)
  
  # Get the initial point with get_initial_design(). This is set to be the point giving the
  # maximum (worst) function evaluation among eval_points
  initial_x, initial_y = utils.get_initial_design(f)
  
  # Initialize GP hyper-parameters. We pre-compute the RBF kernel hyper-parameters
  # for each given f. These are returned by get_hyperparameters().
  hp = utils.get_hyperparameters(f)

  # Initialize kernel and model.
  kernel = GPy.kern.RBF(input_dim=input_dim, variance=hp['variance'],
  lengthscale=hp['lengthscale'], ARD=hp['ard'])
  model = GPy.models.GPRegression(initial_x, initial_y, kernel)

  # Get initial predictive mean and var.
  predictive_mean, predictive_var = model.predict(eval_points)

  # Get initial optimum value.
  found_min = initial_min_y = float(np.min(model.Y))
  
  # Get true optimum value.
  true_min = np.min(f(eval_points))

  # Optimization loop.
  for _ in range(num_trials):
    new_input = acquisition_function(eval_points,  # Get new point using AF. 
        predictive_mean, predictive_var, found_min)
    new_output = f(new_input)  # Evaluate new point.
    model.set_XY(np.concatenate((model.X, new_input), axis=0), # Append to dataset.
                 np.concatenate((model.Y, new_output), axis=0))
    # Get updated mean and var
    predictive_mean, predictive_var = model.predict(eval_points)
    found_min = float(np.min(model.Y))  # Get current optimum value.

  # Get percentage of trials (note that we remove the number of given points in the
  initial design) needed to identify the optimum.
  percentage_steps_before_converging = (np.argmin(model.Y) - len(
    initial_design_inputs)) / (num_trials) if found_min == true_min else 1.0
  return (found_min, true_min, initial_min_y, percentage_steps_before_converging)

\end{python}
\caption{Python code for the first part of $\eval$ used in \funbo. This function runs a full \bo loop with a given number of trials and points on a Sobol grid to assess how efficiently a given \af allows optimizing $f$.}
\label{fig:eval_function_funbo}
\end{figure}

\begin{figure}
\begin{python}
"""Score an AF given the output of run_bo()."""

import numpy as np

def score(found_min, true_min, initial_min_y, percentage_steps_before_converging):
  """Compute a score based on the output of run_bo()."""
    
    # Get score based on how close the found optimum is to the true one (first term
    # in Eq. (1)).
    score_min_reached = 1.0 - np.abs(found_min - true_min) / (initial_min_y - true_min)
    
    # Get score based on how the percentage of trials needed to identify the true
    # optimum (second term in Eq. (1)).
    score_steps_needed = 1.0 - percentage_steps_needed
    
  return score_min_reached + score_steps_needed

\end{python}
\caption{Python code for the second part of $\eval$ used in \funbo. Based on the output of run\_bo(), this function computes a score capturing how accurately and quickly an \af allows identifying the true optimum.}
\label{fig:score_function_funbo}
\end{figure}

\section{Programs Database}\label{sec:programs_database_appendix}
The \db structure matches the one proposed by \funsearch\cite{romera2023mathematical}. We discuss it here for completeness.
A multiple-deme model \cite{tanese1989distributed} is employed to preserve and encourage diversity in the generated programs. Specifically, the program population in \db is divided into $N_{\db}$ islands, each initialized with the given initial $\evolve$ and evolved independently. Within each island, programs are clustered based on their scores on the functions in $\ftrain$, with \af{s} having the same scores grouped together. Sampling from \db involves first uniformly selecting an island and then sampling two \af{s} from it. Within the chosen island, a cluster is sampled, favoring those with higher score values, followed by sampling a program within that cluster, favoring shorter ones. The newly generated \af is added to the same island associated with the instances in the prompt, but to a cluster reflecting its scores on $\ftrain$. Every 4 hours, all programs from the $N_{\db}/2$ islands with the lowest-scoring best \af are discarded. These islands are then reseeded with a single program from the surviving islands. This procedure eliminates under-performing \af{s}, creating space for more promising programs. See the Methods section in \citep{romera2023mathematical} for further details.

\section{Experimental details}
In this section, we provide the experimental details for all our experiments. We run \funbo with $\Tau=48 \text{hrs}$, $B=12$ and $N_{\db}=10$. To isolate the effect of using different \af{s} and eliminate confounding factors related to \af maximization or surrogate models, we maximized all \af{s} on a fixed Sobol grid (of size $\sizesobol$) over each function's input space. We also ensure the same initial design across all methods (including the point with the highest/worst function value on the Sobol grid) and used consistent \gp hyperparameters which are tuned offline and fixed. In particular, we use a \gp model with zero mean function and \acro{rbf} kernel defined as $K_{\theta}(\X, \X') = \sigma^2_f \text{exp}(-||\X-\X'||^2/2\ell^2)$ with $\theta= (\ell, \sigma^2_f)$ where $\ell$ and $\sigma^2_f$ are the length-scale and kernel variance respectively. The Gaussian likelihood noise $\sigma^2$ is set to $1\mathrm{e}-5$ unless otherwise stated. We set $T=30$ for all experiments apart for \expwithinhpo and \expwithingps for which we use $T=20$ to ensure faster evaluations of generated \af{s}. We used the \metabo implementation provided by the authors at \texttt{https://github.com/boschresearch/MetaBO}, retaining default parameters except for removing the local maximization of \af{s} and ensuring consistency in the initial design.  We followed the same procedure for \fsaf, using code available at \texttt{https://github.com/pinghsieh/FSAF}.  We ran \ucb with $\beta=1$. Experiment-specific settings are detailed below.

\subsection{\expacrosssynthetic}\label{sec:across_class_experiments}
\begin{figure}
    \centering
    \includegraphics[width=0.49\textwidth]{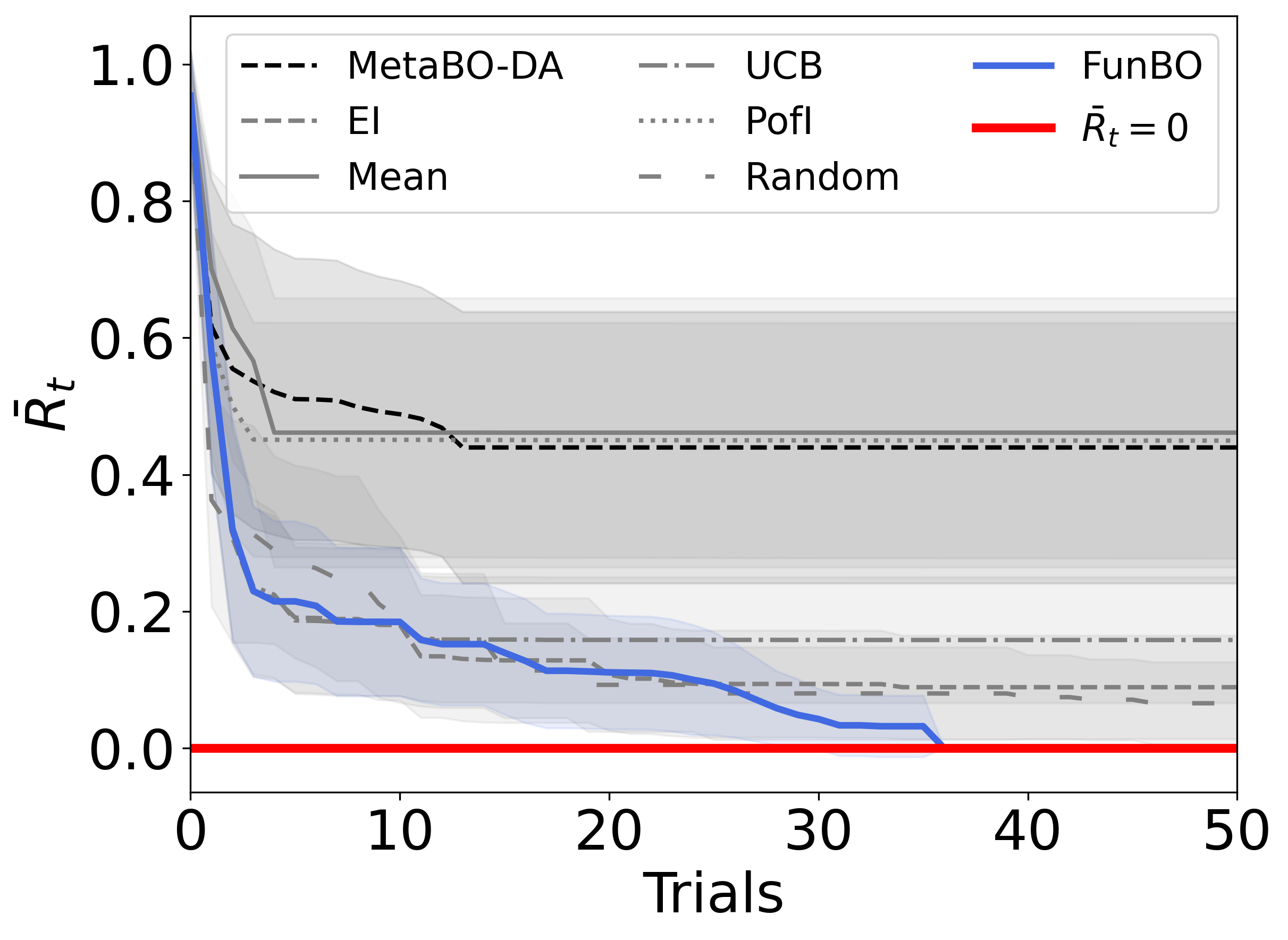}
    \includegraphics[width=0.49\textwidth]{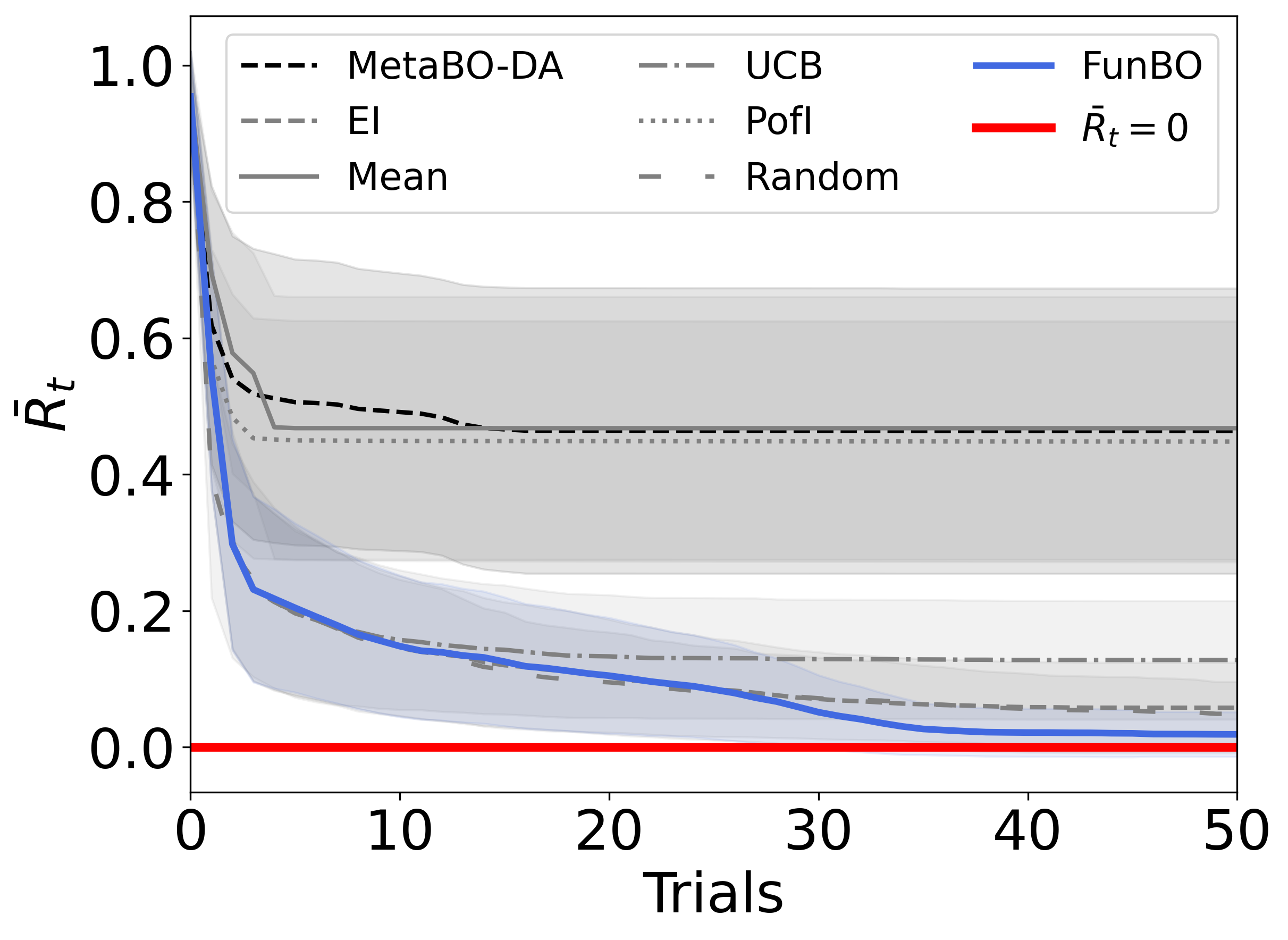}
    \caption{\expacrosssynthetic. Average \bo performance when using known general purpose \af{s} (gray lines with different patterns), the \af learned by a dimensionality agnostic version of \metabo (\metabo-\acro{da}, black dashed line) and $\resultfunbo$ (blue line). Shaded area gives $\pm$ standard deviations$/2$. The red line represents $\bar{R}_t=0$, i.e., zero average regret. \textit{Left}: $\ftest$ includes nine different synthetic functions. \textit{Right}: Extended test set including, for each function in $\ftest$, 50 randomly scaled and translated instances.}
    \label{fig:across_classes_performance_appendix}
\end{figure}

The parameter configurations adopted for each objective function used in this experiment, either in $\fall$ or in $\ftest$, are given in Table \ref{parameters_synthetic_od}. Notice that for \hart with $d=3$ we use an \acro{ard} kernel.
\begin{table}[ht]
\caption{Parameters used for \expacrosssynthetic.}
\label{parameters_synthetic_od}
\resizebox{\columnwidth}{!}{
\begin{tabular}{lcccccccc}
\toprule
 & $d$ & $\mathcal{X}$ & $N_{\acro{sg}}$ & $\ell$ & $\sigma^2_f$ & $\sigma^2_f$ \\
\midrule
Ackley  &  1 & $[-4, 4]$ & 1000 & 0.21 & 28.19 & $1\mathrm{e}-5$\\
Levy   &  1 & $[-10, 10]$ & 1000 & 1.05 & 83.32 & $1\mathrm{e}-5$\\
Schwefel &  1 & $[-500, 500]$ & 1000 & 18.46 & 76868.65 & $1\mathrm{e}-5$ \\
Rosenbrock &  1 & $[-5, 10]$ & 1000 & 1.20 & 87328.20 & $1\mathrm{e}-5$ \\
Sphere &  1 & $[-5, 5]$ & 1000 & 18.46 & 924202.43 & $1\mathrm{e}-5$ \\
\sty &  1 & $[-5, 5]$ & 1000 & 7.34 & 119522207.86 & $1\mathrm{e}-5$ \\
Weierstrass &  1 & $[-0.5, 0.5]$& 1000 & 0.01 & 0.39 & $1\mathrm{e}-5$\\
Beale &  2 & $[-4, 5]^2$ & 10000 & 0.46 & 546837.32 & $1\mathrm{e}-5$ \\
Branin &  2 & $[-5, 10] \times [0, 15]$ & 10000 & 4.65 & 155233.52 & $1\mathrm{e}-5$ \\
Michalewicz &  2 & $[0, \pi]^2$ & 10000 & 0.22 & 0.10 & $1\mathrm{e}-5$\\
\gprice &  2 & $[-2, 2]^2$ & 10000 & 0.27 & 117903.96 & $1\mathrm{e}-5$ \\
\hart-3 &  3 & $[0, 1]^3$ & 1728 & $[0.716,0.298,0.186]$ & 0.83 & $1.688\mathrm{e}-11$\\
\hart-6 &  6 & $[0, 1]^6$ & 729 & 1.0 & 1.0 & $1\mathrm{e}-5$\\
\bottomrule
\end{tabular}}
\end{table}
Scaled and translated functions are obtained with translations sampled uniformly in $[-0.1, 0.1]^\inputdim$ and scalings sampled uniformly in $[0.9, 1.1]$. Fig. \ref{fig:across_classes_performance_appendix} gives the results achieved by $\resultfunbo$ (blue line) and a dimensionality agnostic version of \metabo that does not take the possible evaluation points as input of the neural \af. This allows the neural \af to be trained on one-dimensional functions and be used to optimize functions across input dimensions.

\subsection{\expwithinsynthetic}\label{sec:app_expwithinsynthetic}
\begin{figure}
\centering
\includegraphics[width=0.325\textwidth]{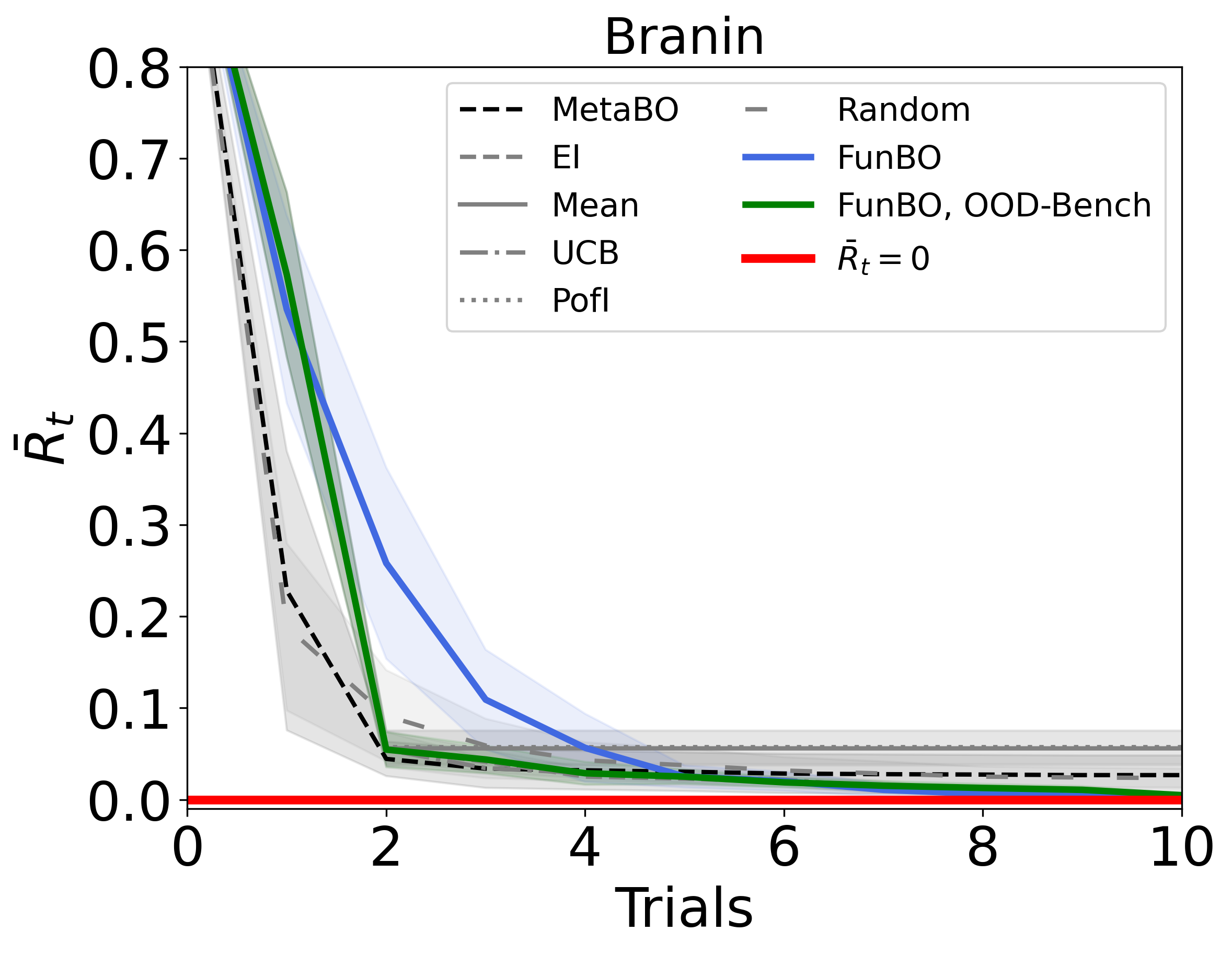}
\includegraphics[width=0.325\textwidth]{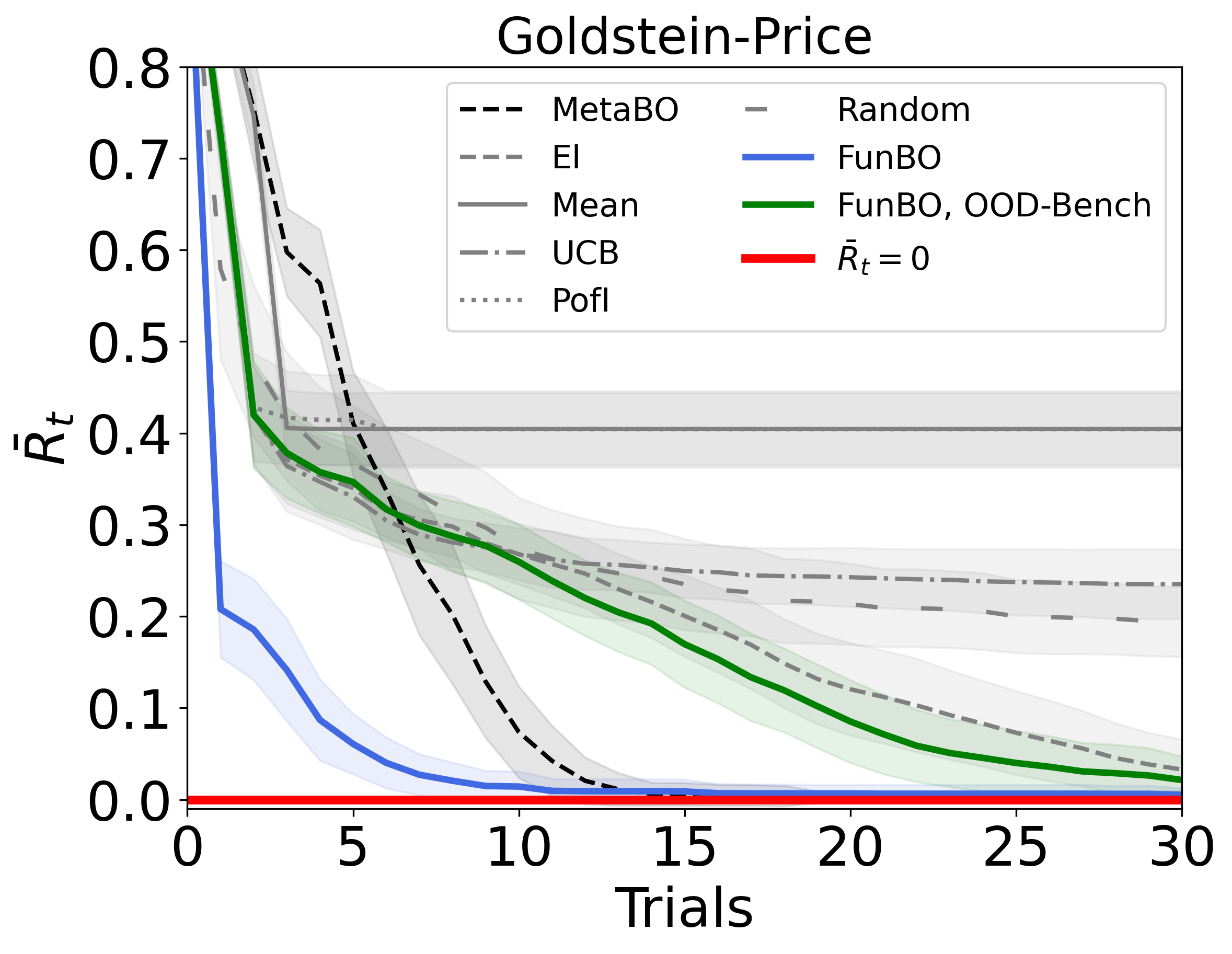}
\includegraphics[width=0.325\textwidth]{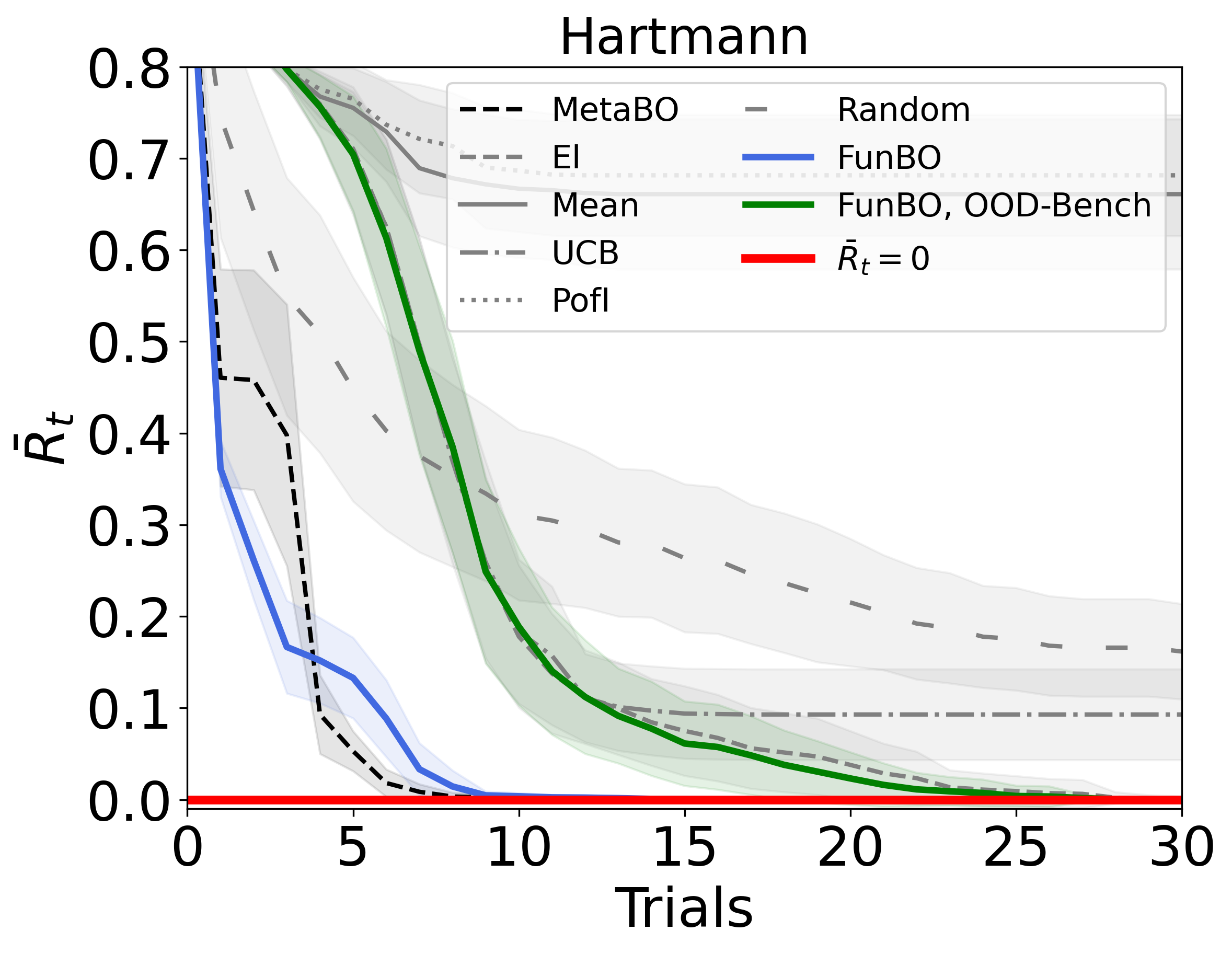}
\caption{\expwithinsynthetic. Average \bo performance when using known general purpose \af{s} (gray lines with different patterns), $\resultfunbo$ found in \expacrosssynthetic (green line), the \af learned by \metabo (black dashed line) and $\resultfunbo$ (blue line) on 100 instances of Branin, \gprice and \hart. Shaded area gives $\pm$ standard deviations$/2$. The red line represents $\bar{R}_t=0$, i.e., zero average regret.}
\label{fig:branin_gprice_hm3_with1D_appendix}
\end{figure}

The parameter configurations for Branin, \gprice and \hart are given in Table \ref{parameters_synthetic_id}. For this experiment, we adopt the parameters used by \citet{volpp2019meta} thus optimize the functions in the unit-hypercube and use \acro{ard} \acro{rbf} kernels. Fig. \ref{fig:branin_gprice_hm3_with1D_appendix} gives the results achieved by $\resultfunbo$ (blue line) and the \af found by \funbo for \expacrosssynthetic (green). The Python code for the found \af{s} is given in Figs. \ref{fig:af_branin}-\ref{fig:af_hm3}.

\begin{table}[ht]
\caption{Parameters used for \expwithinsynthetic.}
\label{parameters_synthetic_id}
\resizebox{\columnwidth}{!}{
\begin{tabular}{lcccccccc}
\toprule
 & $d$ & $\mathcal{X}$ & $N_{\acro{sg}}$ & $\ell$ & $\sigma^2_f$ & $\sigma^2_f$ \\
\midrule
Branin &  2 & $[0, 1]^2$ & 961 & $[0.235, 0.578]$ & 2.0 & $8.9\mathrm{e}-16$ \\
\gprice &  2 & $[0, 1]^2$ & 961 & $[0.130, 0.07]$ & 0.616 & $1\mathrm{e}-6$ \\
\hart-3 &  3 & $[0, 1]^3$ & 1728 & $[0.716,0.298,0.186]$ & 0.83 & $1.688\mathrm{e}-11$\\
\bottomrule
\end{tabular}}
\end{table}

\begin{figure}
\begin{python}
import numpy as np
from scipy import stats

def acquisition_function(predictive_mean, predictive_var, incumbent, beta=1.0):
  """Returns the index of the point to collect ..."""
  y_pred = predictive_mean + 2 * predictive_var
  diff_current_best_y_pred = incumbent - y_pred
  bound_standard_deviation = np.maximum(np.sqrt(predictive_var), 1e-15)
  z = diff_current_best_y_pred / bound_standard_deviation
  vals = (diff_current_best_y_pred * stats.norm.cdf(z)
          + np.sqrt(predictive_var) * stats.norm.cdf(z + 0.5)
          + (stats.norm.cdf(z) - stats.norm.cdf(z + 0.5)) * predictive_var / 2)
  a = np.maximum(diff_current_best_y_pred, incumbent)
  alpha = diff_current_best_y_pred if incumbent > 0.0 else -np.inf
  alpha = np.maximum(alpha, 0.) * (-alpha + 0.5 * a) - y_pred
  y_vals = np.absolute(alpha + a + np.abs(y_pred)) * (a >= 0.)
  for y_val in y_vals:
    idx = np.argmax(vals - (y_val - y_pred) / bound_standard_deviation)
    vals[idx] = 0
  return np.argmax(vals)
\end{python}
\caption{\expwithinsynthetic. Python code for $\resultfunbo$ for Branin. The \bo performance corresponding to this \af is given in Fig. \ref{fig:branin_gprice_hm3} (left).}
\label{fig:af_branin}
\end{figure}

\begin{figure}
\begin{python}
import numpy as np
from scipy import stats

def acquisition_function(predictive_mean, predictive_var, incumbent, beta=1.0):
  """Returns the index of the point to collect ..."""
  shape, dim = predictive_mean.shape
  best_score = 0.0
  g_i = 0.0

  predictive_var[(shape-10)//2] *= dim
  predictive_var[~ np.isfinite(predictive_var)] = 1.0

  for i in range(predictive_mean.shape[0]):
    curr_z = (incumbent - predictive_mean[i]) / np.sqrt(predictive_var[i])
    new_score = predictive_var[i] *  stats.norm.cdf(curr_z, 0.5)
    if new_score > best_score:
      best_score = new_score
      g_i = i
  return g_i
\end{python}
\caption{\expwithinsynthetic. Python code for $\resultfunbo$ for \gprice. The \bo performance corresponding to this \af is given in Fig. \ref{fig:branin_gprice_hm3} (middle).}
\label{fig:af_gprice}
\end{figure}

\begin{figure}
\begin{python}
import numpy as np
from scipy import stats

def acquisition_function(predictive_mean, predictive_var, incumbent, beta=1.0):
  """Returns the index of the point to collect ..."""
  diff_current_best_mean = incumbent - predictive_mean
  standard_deviation = np.sqrt(predictive_var)
  z = diff_current_best_mean / standard_deviation
  vals = diff_current_best_mean * stats.norm.cdf(z)**3 + (
      stats.norm.cdf(z)**2 + stats.norm.cdf(z) + 1) * stats.norm.pdf(z)
  index = np.argmax(stats.truncnorm.cdf(vals, a=-0.1, b=0.1))
  return index
\end{python}
\caption{\expwithinsynthetic. Python code for $\resultfunbo$ for \hart. The \bo performance corresponding to this \af is given in Fig. \ref{fig:branin_gprice_hm3} (right).} 
\label{fig:af_hm3}
\end{figure}

\subsection{\expwithinhpo}\label{sec:app_expwithinhpo}
\begin{figure}
\centering
\includegraphics[width=0.49\textwidth]{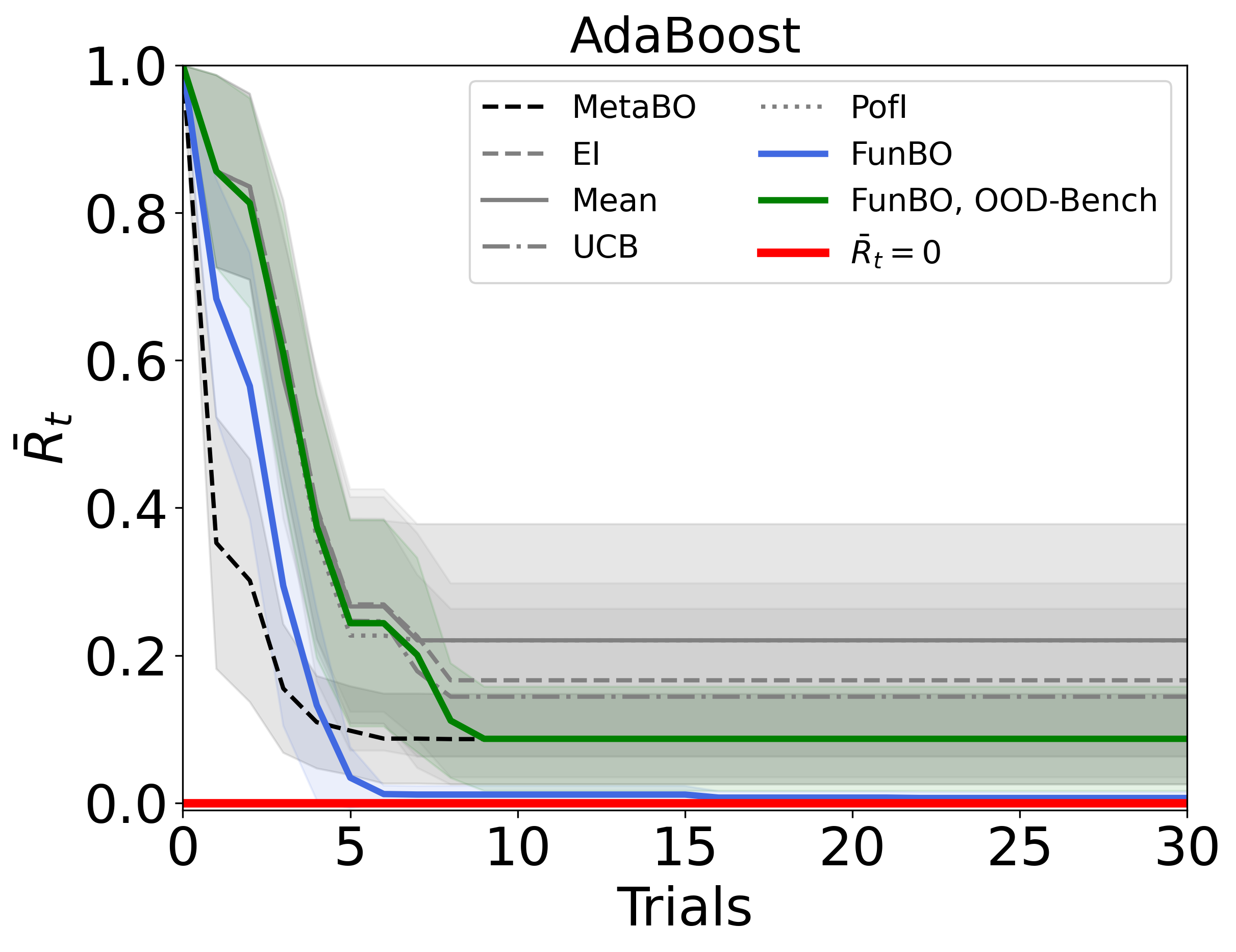}
\includegraphics[width=0.49\textwidth]{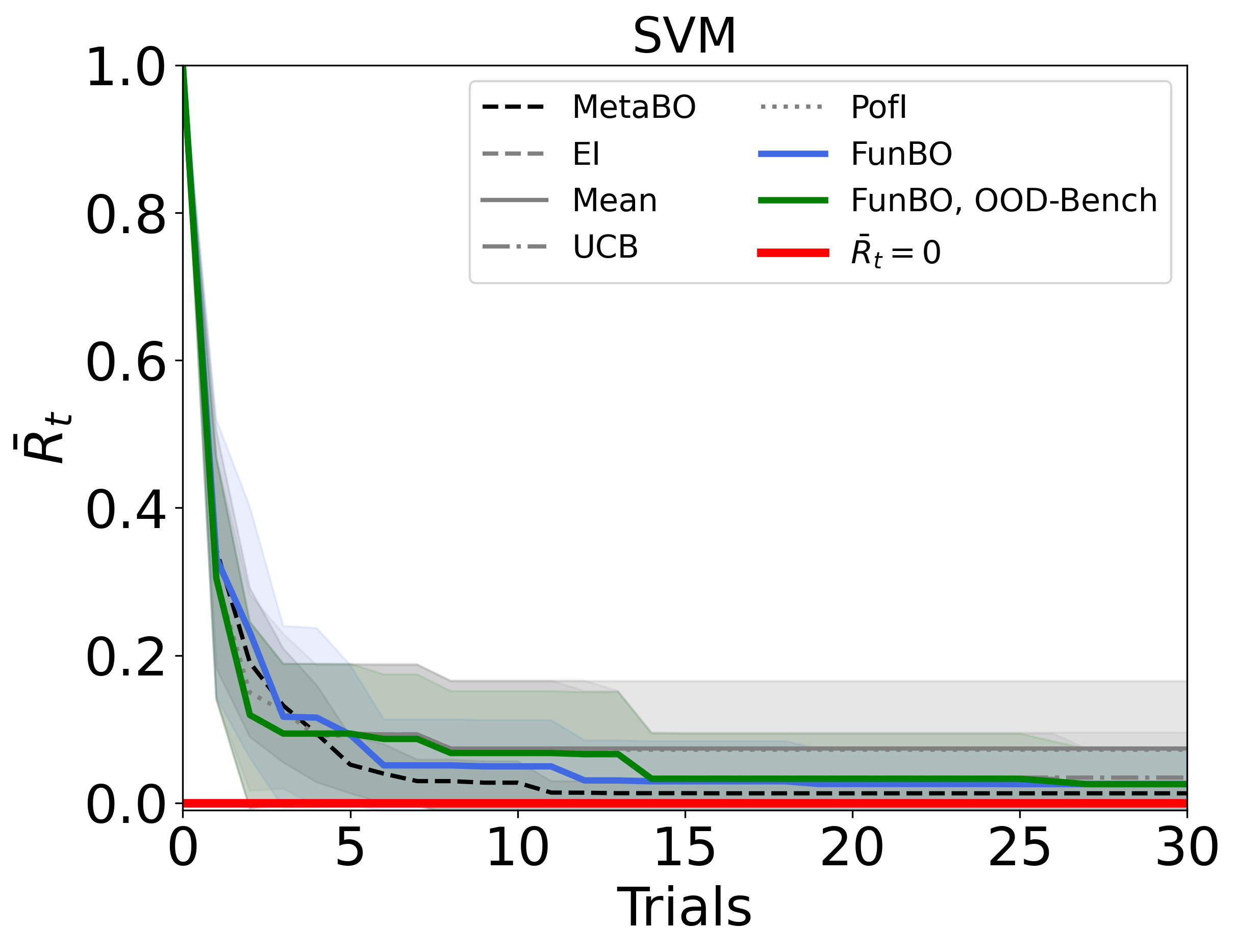}
\caption{\expwithinhpo. Average \bo performance when using known general purpose \af{s} (gray lines with different patterns), a meta-learned \af by \metabo (black dashed line), $\resultfunbo$ found in \expacrosssynthetic (green lines) and $\resultfunbo$ (blue lines). Shaded area gives $\pm$ standard deviations$/2$. The red line represents $\bar{R}_t=0$, i.e., zero average regret.}
\label{fig:adaboost_svm_appendix}
\end{figure}

For this experiment, we adopt the \gp hyperparameters used by \citet{volpp2019meta}. From the training datasets used in \metabo, we assign ``bands'', ``wine'', ``coil2000'', ``winequality-red'' and ``titanic'' for Adaboost, and ``bands'', ``breast-cancer'', ``banana'', ``yeast'' and ``vehicle' for \svm to $\fvalidate$. We keep the rest in $\ftrain$. Fig. \ref{fig:adaboost_svm_appendix} gives the results achieved by $\resultfunbo$ (blue lines) and the \af found by \funbo for \expacrosssynthetic (green lines). The Python code for the found \af{s} is given in Figs. \ref{fig:af_adaboost}-\ref{fig:af_svm}.

\begin{figure}
\begin{python}
import numpy as np
from scipy import stats

def acquisition_function(predictive_mean, predictive_var, incumbent, beta=1.0):
  """Returns the index of the point to collect ..."""
  c1 = np.exp(-beta)
  c2 = 2.0 * beta * np.exp(-beta)
  alpha = np.sqrt(2.0) * beta * np.sqrt(predictive_var)
  z = (incumbent - predictive_mean) / alpha
  vals = -abs(c1 * np.exp( - np.power(z, 2)) - 1.0 + c1 + incumbent
    ) + 2.0 * beta * np.power(z+c2, 2)
  vals -= np.log(np.power(alpha, 2))
  vals[np.argmin(vals)] = 1.0
  return np.argmin(vals)
\end{python}
\caption{\expwithinhpo. Python code for $\resultfunbo$ for \adaboost. The \bo performance corresponding to this \af is given in Fig. \ref{fig:adaboost_gps} (left).}
\label{fig:af_adaboost}
\end{figure}

\begin{figure}
\begin{python}
import numpy as np
from scipy import stats

def acquisition_function(predictive_mean, predictive_var, incumbent, beta=1.0):
  """Returns the index of the point to collect ..."""
  z = (incumbent - predictive_mean) / np.sqrt(predictive_var)
  vals = (incumbent - predictive_mean) * stats.norm.cdf(z
    ) + np.sqrt(predictive_var) * stats.norm.pdf(z)
  t0_val = stats.norm(loc=incumbent, scale=np.sqrt(predictive_var)).pdf(incumbent)
  t1_val = z * stats.norm.pdf(z)
  vals = ((vals * t1_val - t0_val) / (1 - 2 * t1_val)
          + t1_val*(vals/(1-2*t1_val)) 
          - vals/(1 - 2*t1_val)**2 + t1_val*(t1_val - z)/beta)
  return np.argmax(vals)
\end{python}
\caption{\expwithinhpo. Python code for $\resultfunbo$ for \svm. The \bo performance corresponding to this \af is given in Fig. \ref{fig:adaboost_svm_appendix} (right).}
\label{fig:af_svm}
\end{figure}

\subsection{\expwithingps}\label{sec:app_expwithingps}
\begin{figure}
    \centering
    \includegraphics[width=0.49\textwidth]{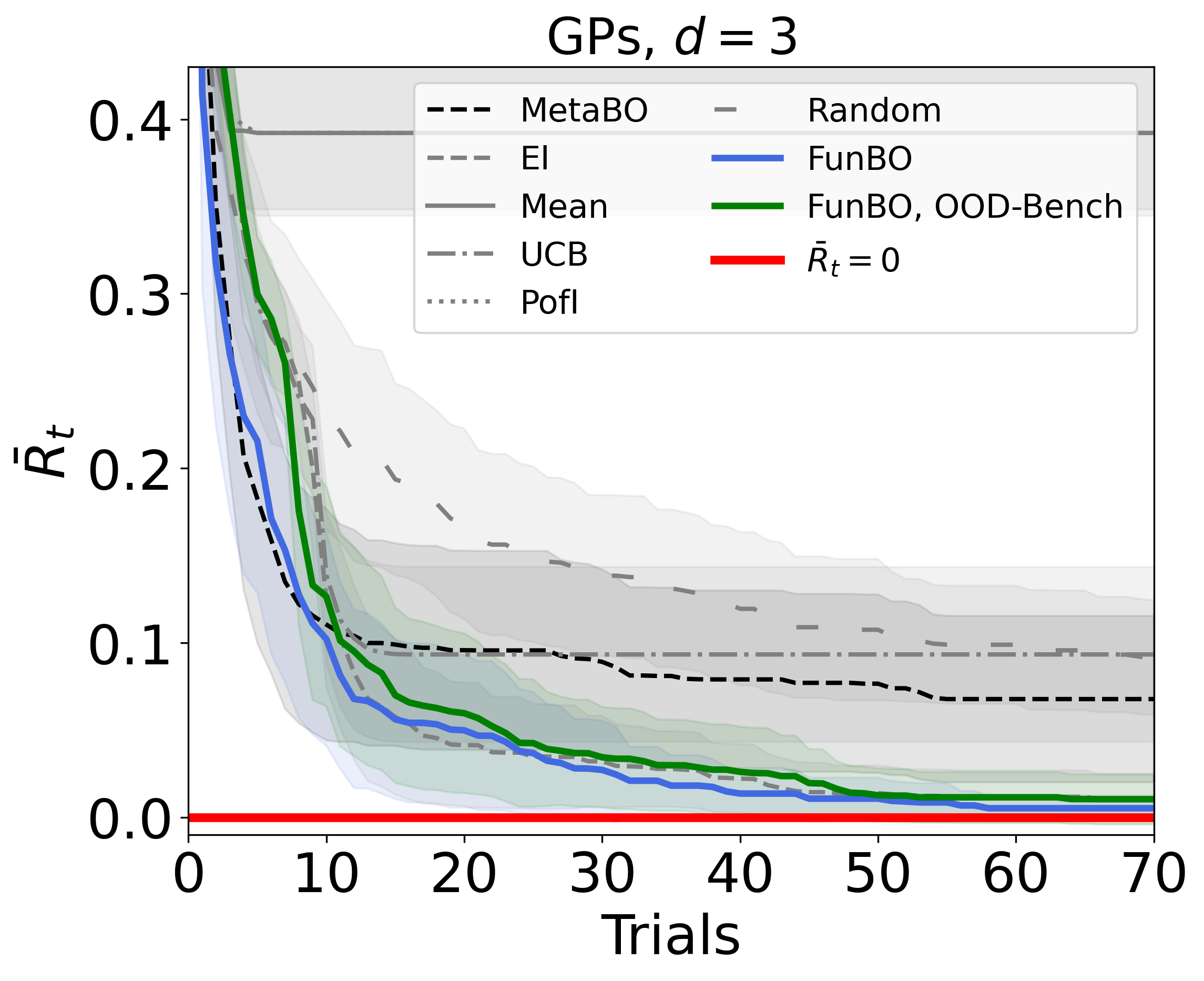}
    \includegraphics[width=0.49\textwidth]{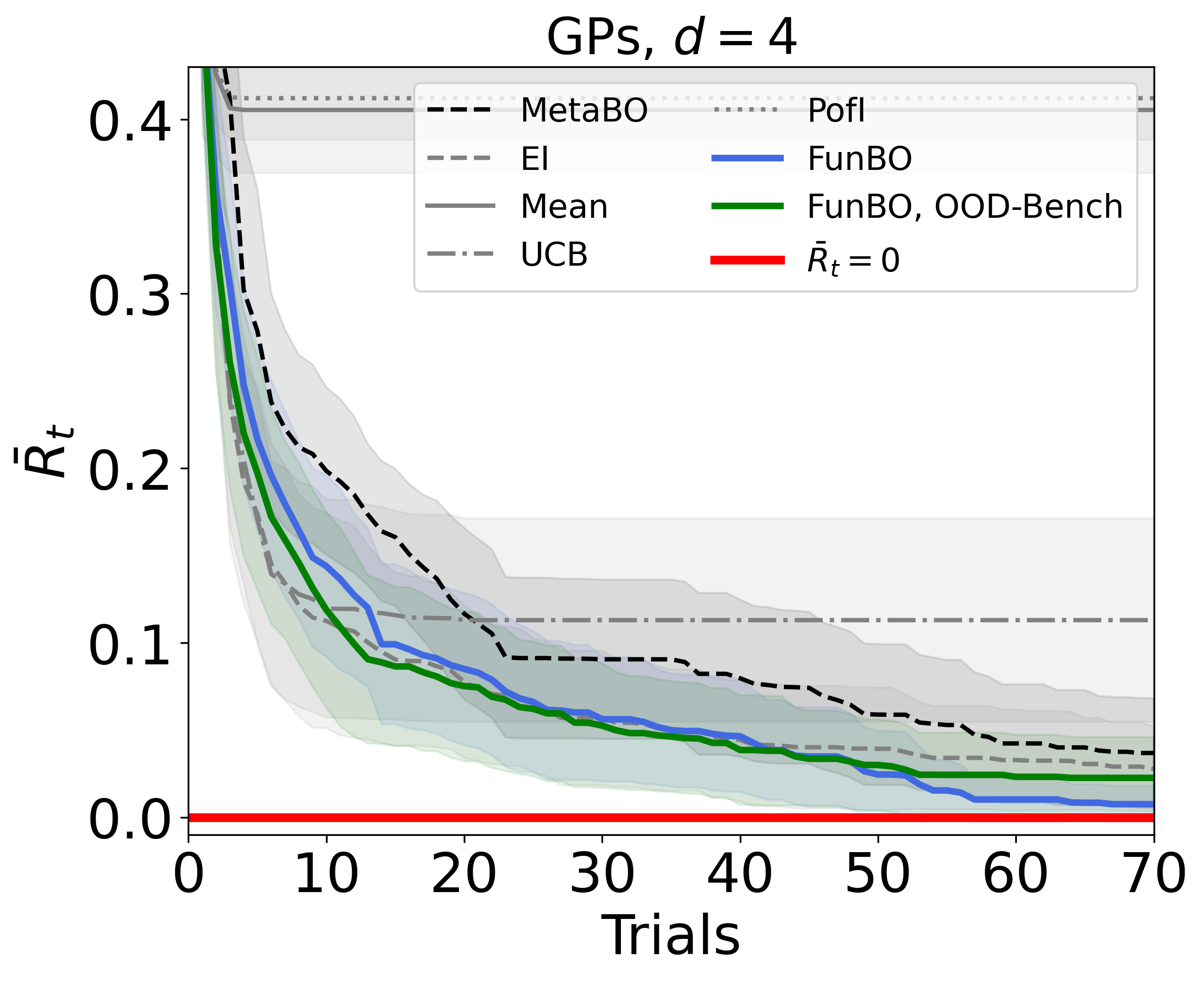}
    \caption{Average \bo performance  when using known general purpose \af{s} (gray lines with different patterns), the \af learned by \metabo (black dashed line), $\resultfunbo$ found in \expacrosssynthetic (green lines) and $\resultfunbo$ (blue lines). Shaded area gives $\pm$ standard deviations$/2$. The red line represents $\bar{R}_t=0$, i.e. zero average regret. \textit{Left}: \expwithingps. $\ftest$ includes functions with $\inputdim=3$. \textit{Right}: $\ftest$ includes functions with $\inputdim=4$.}
    \label{fig:gps_appendix}
\end{figure}

The functions included in both $\fall$ and $\ftest$ are sampled from a \gp prior with \acro{rbf} kernel and length-scale values drawn uniformly from $[0.05, 0.5]$. The functions are optimized in the input space $[0, 1]^3$ with $N_{\acro{sg}}=1728$ points. In terms of \gp hyperparameters, we set $\sigma^2_f=1.0$, $\sigma^2=1\mathrm{e}-20$ and use the length-scale value used to sample each function as $\ell$. Fig. \ref{fig:gps_appendix} gives the results achieved by $\resultfunbo$ and the \af found by \funbo for \expacrosssynthetic. The Python code for $\resultfunbo$ is given in Fig. \ref{fig:af_gps}.

\begin{figure}
\begin{python}
import numpy as np
from scipy import stats

def acquisition_function(predictive_mean, predictive_var, incumbent, beta = 1.0):
  """Returns the index of the point to collect ..."""
  z = (incumbent - predictive_mean) / np.sqrt(predictive_var)
  vals = ((incumbent - predictive_mean) * stats.norm.cdf(z
    ) + np.sqrt(predictive_var) * stats.norm.pdf(z))**2
  vals = vals / (1 + (z / beta)**2 * np.sqrt(predictive_var))**2
  return np.argmax(vals)
\end{python}
\caption{\expwithingps. Python code for $\resultfunbo$. The \bo performance corresponding to this \af is given in Fig. \ref{fig:adaboost_gps} (right).}
\label{fig:af_gps}
\end{figure}

\subsection{\fewshot}\label{sec:app_fewshot}
For this experiment, the 5 Ackley functions used to ``adapt'' the initial \af are obtained by scaling and translating the output and inputs values with translations and scalings uniformly sampled in $[-0.1,0.1]^d$ and $[0.9,1.1]$ respectively. The test set includes 100 instances of Ackley similarly obtained with scale and translations values in $[0.7,1.3]$ and $[-0.3,0.3]^d$ respectively. Furthermore, we consider $[0, 1]^2$ as input space and use $N_{\acro{sb}}=1000$. The \gp hyperparameters are set to $\ell=[0.07, 0.018]$ (\acro{ard} kernel), $\sigma^2_f = 1.0$ and $\sigma^2 = 8.9\mathrm{e}-16$. Python code for $\resultfunbo$ is given in Fig. \ref{fig:af_fewshots}.

\begin{figure}
\begin{python}
import numpy as np
from scipy import stats

def acquisition_function(predictive_mean, predictive_var, incumbent, beta=1.0):
  """Returns the index of the point to collect ..."""
    num_points, _ = predictive_mean.shape
  a = 10
  z = (predictive_mean + 0.000001 - incumbent) / np.sqrt(predictive_var)
  vals = 1 / ((1 + (z / beta)**2 * np.sqrt(a * predictive_var + 0.00001)) **2)
  beta_sqrt_p_z = np.sqrt(beta) * z
  vals *= (1 + (z / beta)**2)*predictive_var/(
      (1+ (beta_sqrt_p_z / np.sqrt(predictive_var))**2 * predictive_var) * (
          1+(beta_sqrt_p_z / np.sqrt(predictive_var))**2))
  vals += (1 - beta_sqrt_p_z / np.sqrt(predictive_var))**2 * predictive_var/ (
      1 + (beta_sqrt_p_z / np.sqrt(predictive_var))**2 * predictive_var)**2
  vals = (1 + (z / beta)**2) * vals- (1 - (z / beta)**2) * np.exp(- 1) ** 2
  vals = np.sqrt(a * predictive_var) * vals / np.sqrt(
      a * predictive_var + 0.00001)
  vals *= np.sqrt(np.sqrt(a * predictive_var) * predictive_var)
  vals *= predictive_var**2
  vals[:num_points // 2] = 0
  return np.argmax(vals)
\end{python}
\caption{\fewshot. Python code for $\resultfunbo$. The \bo performance corresponding to this \af is given in Fig. \ref{fig:few_shots}.}
\label{fig:af_fewshots}
\end{figure}


\end{document}